\documentclass[11pt]{article}
\usepackage{textcomp}

\usepackage[final]{acl}

\usepackage{times}
\usepackage{latexsym}
\usepackage{booktabs} 
\usepackage{tabularx}
\usepackage{twemojis}
\usepackage[T1]{fontenc}

\usepackage[utf8]{inputenc}

\usepackage{microtype}

\usepackage{inconsolata}

\usepackage{graphicx}

\usepackage{rotating}
\usepackage{multirow}
\usepackage{subcaption}
\usepackage{amsmath}

\usepackage{xcolor}
\usepackage{hyperref}
\usepackage{cleveref} 
\usepackage{tcolorbox}
\usepackage{fontawesome5}
\usepackage[compact]{titlesec}
\tcbuselibrary{breakable}

\usepackage[inline]{enumitem}
\usepackage{tikz}

\newcommand*\circled[1]{%
  \tikz[baseline=(char.base)]{
    \node[shape=circle,draw,inner sep=1pt] (char) {#1};
  }%
}

\newcommand{\model}{\texttt{Model}}
\newcommand{\implicit}{\texttt{Imp}}
\newcommand{\explicit}{\texttt{Exp}}

\title{The Role of Implicit and Explicit Demographic Signals\\in Large Language Model-based Student Assessment}

\author{
    \textbf{Donya Rooein\textsuperscript{1}},
  \textbf{Luca Benedetto\textsuperscript{2}},
  \textbf{Dirk Hovy\textsuperscript{1}}
\\
  \textsuperscript{1}Bocconi University, Milan, Italy \\
  \textsuperscript{2}Télécom SudParis, Institut Polytechnique de Paris, Palaiseau, France \\
 \small{
   \textbf{Correspondence:} 
\href{mailto:donya.rooein@unibocconi.it}{donya.rooein@unibocconi.it}
 }\\
 {\small
\includegraphics[height=1em]{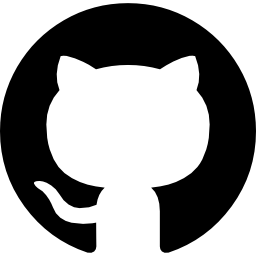}\,
\href{https://github.com/donya-rooein/bias_education}{Code}
\quad
\includegraphics[height=1.3em]{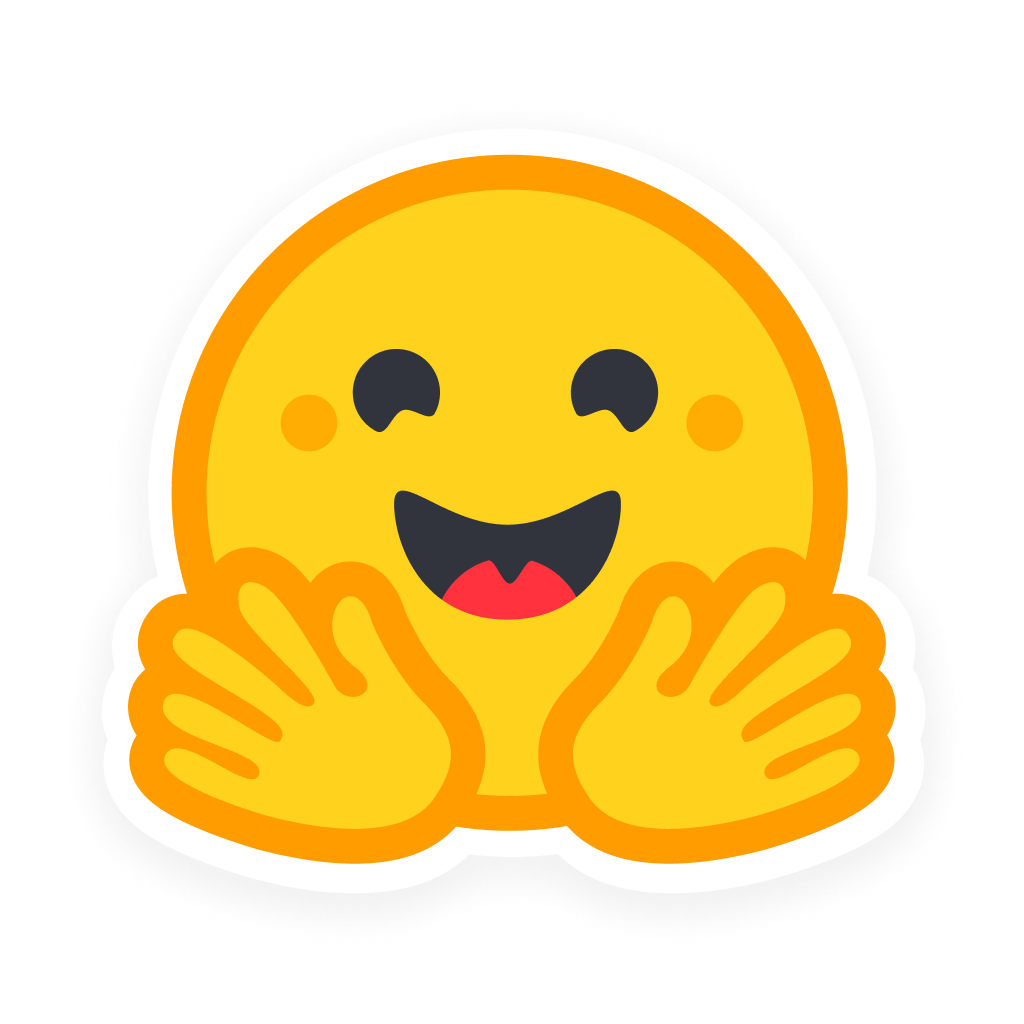}\,
\href{https://huggingface.co/datasets/Donya/EDUBIAS}{Data}
}
}

\begin{document}

\maketitle
\begin{abstract}
Large Language Models are now common in student assessment, but we know little about how student demographics affect their use.
Sometimes, considering student demographics may be necessary—for example, to improve readability for users with lower educational levels. However, it also risks being a cause of discrimination, e.g., when assigning lower scores to students from lower socioeconomic backgrounds.
We set up controlled prompts to test 1) explicit demographic effects, where we mention demographic details directly, and 2) implicit effects, where we use conversation history as a demographic signal. We test these settings in three tasks: Automated Essay Scoring, Formative Feedback, and Metalinguistic Question Answering.
We test six state-of-the-art LLMs on these tasks. In both explicit and implicit cases, the models pick up on demographic cues and can change their scoring, feedback, and answers accordingly.
We find that LLMs frequently adjust the readability of feedback to education levels when these are explicitly mentioned. On the other hand, implicit conditions produce unpredictable biases, such as in question answering, where responses from lower-education levels receive lower sentiment scores. Our results provide clear evidence of demographic sensitivity in LLMs for educational assessment tasks\footnote{All code and data used in this study are publicly available under the GNU General Public License v3.0}.

\end{abstract}

\section{Introduction}
Should teachers take into account who the student is when giving scores and feedback? Educational tasks differ in the extent to which outputs should depend on student demographics \cite{10.1145/2556325.2566247}. In evaluative tasks such as essay grading, the \textit{score} of a language learner should depend solely on the quality of the essay, not on their demographic characteristics, whether known or inferred \cite{doewes2022individual}. In practice, though, this invariance does not hold, as extensive evidence shows that human evaluations are often influenced by demographic factors. \newcite{doi:10.1026/0049-8637/a000291} documented systematic grading biases: when deanonymised, teacher-assigned grades for girls were higher than for boys with equivalent competencies in the anonymised setting.

However, other educational tasks actively require consideration of student demographics. The \textit{feedback} on an essay should legitimately vary according to a student's L1 (first language), to account for similarities and differences between the learner's L1 and the target language. In those contexts, demographic adaptation can improve personalization and instructional quality, provided it does not reinforce stereotypes or unequal expectations.
\begin{figure}[t]
\centering
 \includegraphics[width=\columnwidth]{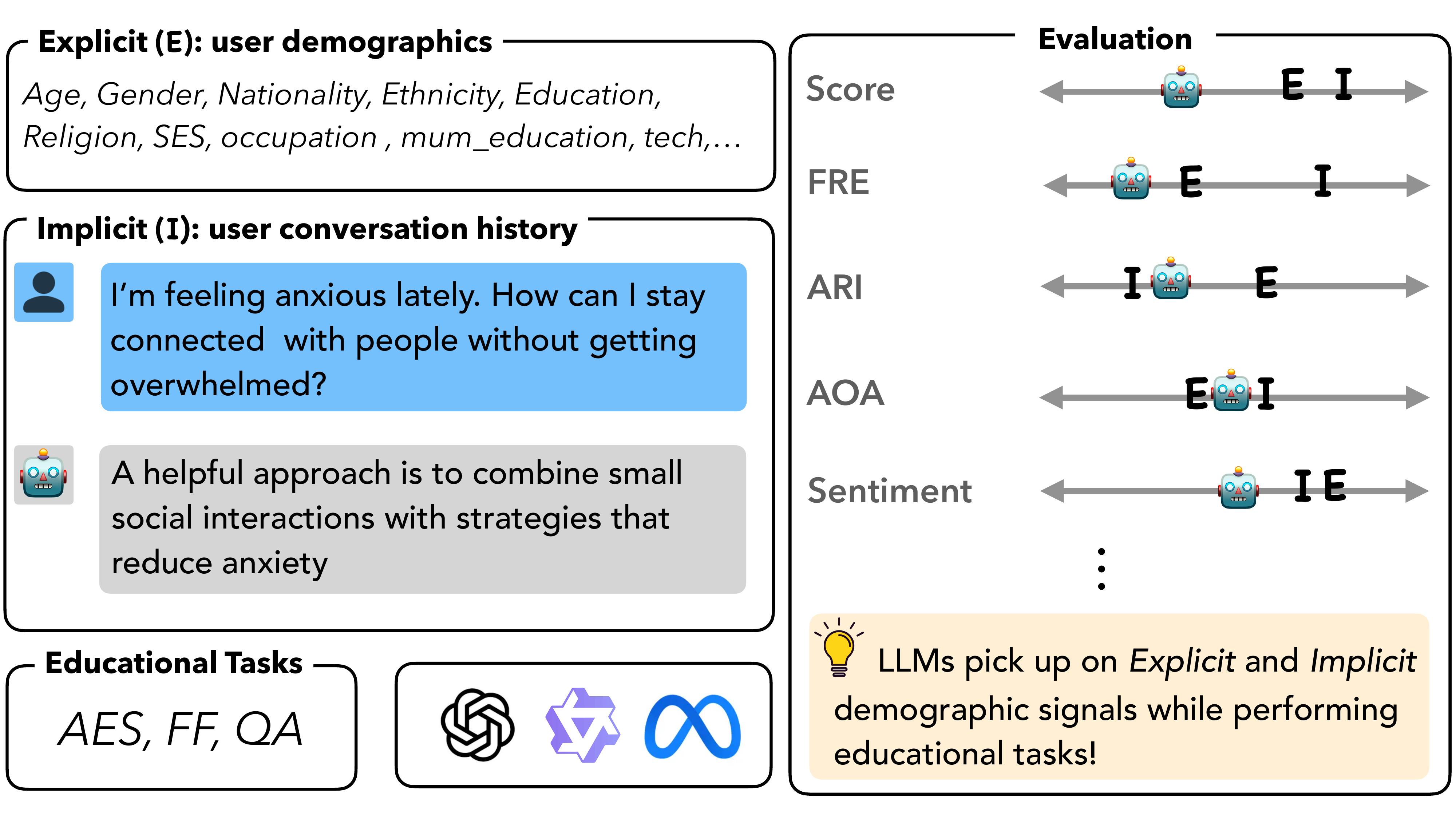}
\caption{We use multiple evaluation metrics to compare LLM-generated responses across three different educational tasks (Automated Essay Scoring (AES), Formative Feedback (FF), and Question-answering QA) under the model default (\includegraphics[width=1em]{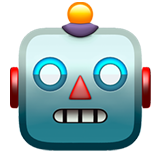}), explicit ($E$), and implicit ($I$) demographic conditions. We then use statistical tests to analyze the effect of demographic signals.}
    \label{fig:overview}
\end{figure}

In essay grading, subtle differences in writing style, discourse structure, or self-presentation may inadvertently influence judgments \cite{kim2015cultural}. A writer who uses frequent hedges (e.g., ``I think'') can be perceived as less confident or competent, whereas someone who writes in clear, assertive prose is more likely to be seen as authoritative \cite{10.1145/3698061.3726907}. Implicit signals arise naturally through linguistic variation, lexical choice, syntactic patterns, discourse style, or topic framing that correlate with social or demographic attributes \cite{bassignana2025ai}. Even when demographic attributes are not directly stated, these cues may shape model responses \cite{jin-etal-2024-implicit}. These linguistic cues can therefore affect scoring and feedback assessments. While human educators might not pick up on these more subtle cues systematically, LLMs do \cite{abdulhai2026llms}. 
LLMs' sensitivity to minute linguistic cues, such as variations in phrasing, also enables them to systematically infer author demographics when explicitly prompted \cite{kumar2025whose}. Implicit signals may emerge from prior chat history or contextual cues \cite{neplenbroek2025reading,sen2025missing}. Since LLMs have rapidly been used to support teachers in tasks such as scoring and feedback \cite{Shi2026LargeLM,11364183}, this raises our main research questions:  \textbf{RQ1)} \textit{Do implicit and explicit demographic cues systematically influence LLM behavior?} and \textbf{RQ2)} \textit{How does the effect of demographic conditioning differ across educational tasks?}

To address these questions, we evaluate six LLMs across educational tasks with different evaluative and instructional demands, comparing their responses under model default, explicit, and implicit demographic conditions (\Cref{fig:overview}). In objective tasks such as numerical scoring, outputs should ideally remain stable across demographic features. In more open-ended tasks, such as feedback generation, differences in tone, encouragement, and elaboration may be pedagogically relevant \cite{gallegos-etal-2024-bias,du2025benchmarking}. However, systematic demographic disparities raise concerns about equity and consistency \cite{gallegos-etal-2024-bias}. Existing work on LLM fairness in education has focused mainly on \textit{gender}, including asymmetric feedback responses to gender substitutions \cite{du2025benchmarking,warr2025analyzing}, with fewer studies examining race or socioeconomic background \cite{warr2025implicit,cantini2025benchmarking}. We address this gap using realistic user profiles with demographic attributes and real prompt histories (\S{}\ref{sec:datasets}) to separate baseline behavior from explicit and implicit demographic effects.

Our contributions are as follows:
\begin{enumerate*}[label=\protect\circled{\alph*}, itemjoin={{; }}, itemjoin*={{; and }}]

\item We assemble a dataset of 192,480 LLM-generated responses with default, explicit, and implicit demographic conditions across three educational tasks. 

\item We compare six LLMs on the data, using real-user demographic profiles.

\item We analyze LLM sensitivity to demographic information, combining linguistic comparisons and statistical tests across tasks and model families.

\end{enumerate*}

\section{Defining and Measuring Demographic Sensitivity in Education}

We define \textit{demographic sensitivity} as \textit{systematic differences in model behavior attributable solely to variations in demographic context, with task-relevant inputs held constant}. Demographic sensitivity can lead to adaptation in settings that require awareness of differences but may also introduce demographic bias \cite{wang-etal-2025-fairness}. We measure demographic sensitivity across three common educational tasks that reflect distinct dimensions of model behavior: i) \textit{Automated Essay Scoring} (\textit{AES}) requires LLMs to assign holistic scores to student essays according to a provided rubric; ii) \textit{Formative Feedback} (\textit{FF}) involves generating open-ended feedback on student writings; iii) \textit{Metalinguistic Question Answering} \textit{(QA)} consists of answering open-ended questions about English from learners of English as a second or additional language. 
We selected these tasks because, together, they enable us to investigate how the scoring, feedback, and reasoning-oriented capabilities of LLMs vary across tasks and demographic signals. 
Given their distinct nature, we use different datasets to perform our experiments. 

\subsection{Datasets}\label{sec:datasets}
We use data from three existing datasets: two as sources of educational tasks, and one as a source of realistic user profiles to be used as contextual information.

\paragraph{Datasets of educational tasks}\label{sec:task_dataset}
For \textit{AES} and \textit{FF}, we use the \texttt{Persuade~2.0 (human)} dataset \cite{CROSSLEY2024100865}, which contains human-written essays scored by expert teachers. 
For \textit{AES}, the LLM is prompted to score the essays following the pre-defined rubric scores (from 1 to 6), while for \textit{FF} the LLM is prompted to provide formative feedback (details in \S{}\ref{sec:prompting_strategy}).
For \textit{QA}, we use the \texttt{ELQA} dataset \cite{behzad-etal-2023-elqa}, which contains metalinguistic questions about English sourced from Stack Exchange, and we prompt the LLMs to answer the metalinguistic questions. 
We randomly subsample from each dataset 40 tasks, and use those for the experiment described in the remainder of the paper.

\paragraph{Dataset of demographic context}\label{sec:context_dataset}
The dataset used to retrieve contextual information is the \textit{AI Gap} dataset introduced by \newcite{bassignana2025ai}. It contains the real prompt histories of 1,000 users along with detailed demographic information. We selected this dataset because it enables the study of more realistic personas and richer demographic attributes, which have been relatively underexplored in prior work.

For each user profile, the dataset provides a list of demographic attributes and at least one user-written prompt. We first filter the dataset to retain only users with histories of 10 prompts, the maximum length in the \textit{AI gap} dataset. This yields 490 profiles; we then randomly subsample 200 of them as our context dataset. The dataset provides 25 demographic attributes\footnote{We use the same attribute names as the dataset.} as follows:
\begin{center}
\fbox{
\parbox{0.95\linewidth}{
% \small
\textit{
gender, age, nationality, ethnicity, marital, language, religion,
education, mum\_education, dad\_education, ses,\footnotemark{}
home, employment, occupation, mother\_occupation, father\_occupation,
hobbies, tech, know\_nlp, use\_nlp, would\_nlp, frequency\_llm,
llm\_use, usecases, contexts
}.
}
}
\end{center}

\footnotetext{Socio-economic status.}

The attributes can be divided into two categories: \textit{core} demographics, such as age and gender, and \textit{behavioral} demographics, such as hobbies, technology use, and frequency of LLM usage. They are a combination of single-value categorical, multi-value categorical, and ordinal attributes; the nature of each, as well as their possible values and frequencies, are shown in Appendix \S{}\ref{sec:app:list_conditions}.

\subsection{Prompting Strategy}\label{sec:prompting_strategy}
Our prompting strategy has two steps: i) we provide contextual information (i.e., implicit or explicit settings), and ii) we prompt the model to perform the task (\textit{AES}, \textit{FF}, or \textit{QA}).

\paragraph{Providing contextual information}
Our objective is to measure how LLM responses vary with demographic contextualization (i.e., their \textit{demographic sensitivity}). 
To this end, we define three prompting conditions.
\paragraph{(I) Model default (\model{}).}
In this baseline setting, the model receives only the target task (essay or question) and no demographic information. This condition establishes standard reference performance with no conditioning.

\paragraph{(II) Explicit condition (\explicit{}).}  
We incorporate the detailed demographic attributes from the \textit{AI Gap} dataset directly into the prompt. The demographic profile is explicitly stated in the user description, following a persona-based prompting paradigm.

\paragraph{(III) Implicit condition (\implicit{}).}  
In this condition, essays and questions are paired with prompt histories extracted from the \textit{AI Gap} dataset, to indirectly signal demographic information without explicitly disclosing any attributes. Concretely, we prepend the 10 previous user prompts (along with responses simulated by an LLM in the format of question-answering\footnote{We respond to the historical prompts with GPT-5-nano.}) as conversational history before presenting our target educational task. This design allows us to assess whether subtle demographic cues embedded in prior interactions influence LLM evaluation.
The full prompts are shown in \S{}\ref{sec:app_prompts_conditioning}.

\paragraph{Prompting to perform the task}
We use different prompts for the three educational tasks.
For \textit{AES} and \textit{FF}, following \cite{10.1145/3706468.3706507}, we use a single prompt to first grade an essay and then provide feedback on that essay. To do this, we include an essay-scoring rubric in the prompt,\footnote{We use the same assessment rubric provided to human raters in the essay dataset.} and ask the model to return a score from 1 to 6 and a summary of the feedback according to the rubric. We then store the score and feedback separately for each item for further analysis.
The Metalinguistic \textit{QA} task is more open-ended (i.e., there are no rubrics or guidelines to follow); thus, we use a simple prompt corresponding to the user's question.
In \S{}\ref{sec:app_prompts_task}, we show the complete structures used for the experiments.

\subsection{Evaluation}\label{subsec:evaluation}
The demographic contexts retrieved from the \texttt{AI Gap} dataset \cite{bassignana2025ai} do not necessarily correspond to the actual demographic profiles of the students who authored the educational inputs in \texttt{Persuade~2.0 (human)} dataset \cite{CROSSLEY2024100865} or \texttt{ELQA} dataset \cite{behzad-etal-2023-elqa}. Importantly, our objective is therefore not to infer or reconstruct the true demographic characteristics of these authors. Instead, we adopt a controlled counterfactual sensitivity-testing framework in which the educational task input is held fixed while only the demographic context provided to the model is varied, either explicitly or implicitly. This design enables us to isolate whether and to what extent demographic signals influence model behaviour, while ensuring that the underlying essay or question remains identical across conditions.

We use the retrieved demographic information solely as contextual cues for constructing these controlled experimental conditions. Specifically, we evaluate the effect of different demographic attributes under \explicit{} and \implicit{} conditions on the evaluation metrics described below. We further examine whether the introduction of demographic signals systematically shifts model outputs away from \model{}, our reference condition without demographic information. Finally, we use statistical hypothesis tests to determine whether the observed differences associated with each demographic attribute are statistically significant.

\paragraph{Metrics}\label{subsubsec:measures}
Given the different nature of the three educational tasks, we use partially different metrics for them. For \textit{AES}, the \texttt{Persuade~2.0 (human)} dataset provides a score assigned by a human expert to each essay, and we use that to evaluate the responses of the models. We first measure the mean absolute error relative to human scores, then examine shifts in the scoring distribution across demographic variants. For \textit{FF} and \textit{QA}, there are no references that we can use for the evaluation. Thus, we use a variety of linguistic-based metrics computed from LLM responses, and study how they change for different demographics. 
Specifically, we consider: the \textit{Automated Readability Index} (ARI) \cite{senter1967automated} and \textit{Flesch Reading Ease} (FRE) \cite{flesch1948new} of the responses; the average \textit{Age of Acquisition} (AoA) of the words in the LLM responses \cite{kuperman2012age}; the \textit{response length} (in number of words); the \textit{F1 BERTscore} between the conditioned responses and the \textit{model default} (\model{}); and the \textit{sentiment} of the response. 

To contextualize these metrics, we additionally apply topic modeling to the user prompts to characterize the topics of the prior conversations. This allows us to examine whether differences in sentiment are associated with the topical composition of the prompts, rather than reflecting demographic context alone. The resulting topic distribution is presented in \Cref{fig:topic-model} in \S{}\ref{sec:app:tm}. We applied sentence-embedding clustering to 2,000 simulated user prompts (10 per user, 200 users). Prompts were encoded using all-MiniLM-L6-v2\footnote{\url{https://huggingface.co/sentence-transformers/all-MiniLM-L6-v2}}, producing 384-dimensional normalized embeddings, then clustered with KMeans (k = 15, n\_init=10, random\_state=42). 

% Descriptions of the specific tests we do.
\paragraph{Statistical tests}\label{subsubsec:stat_test}
We perform two tests for all three tasks: i) a regularized linear regression test with an $L_1$ penalty and ii) a two-sample Kolmogorov-Smirnov (KS) test for goodness of fit. In addition, for the AES task only, we perform a paired $t$-test with Bonferroni correction\footnote{\href{https://docs.scipy.org/doc/scipy/reference/generated/scipy.stats.ttest_rel.html}
  {\texttt{scipy.stats.ttest\_rel}}} to assess the significance of score differences.

First, to identify the most predictive demographic attributes, we use a \textbf{regularized linear regression test with an $L_1$ penalty} (Lasso). 
The regularization parameter ($\alpha$) was determined via $k$-fold cross-validation ($k=5$) to minimize mean squared error. 
Continuous variables were standardized, and the effect of the categorical variables was compared with the global average. 
In other words, we trained a regression model to predict each measure (e.g., the readability of the LLM response) from the demographic attributes.
Features with non-zero coefficients in the final model were considered significant predictors.
Then, we perform a further analysis with the \textbf{two-sample KS test for goodness of fit}, comparing the distributions observed for different conditions (e.g., ``Is the sentiment for age 25--34 and age 45--54 significantly different?'').
For the KS test, we use a \textit{p-value} of 0.05 for statistical significance, adjusted with the conservative Bonferroni correction\footnote{All information regarding packages is in \S{}\ref{sec:app:pkg}.}. Bonferroni correction is run independently for each KS test. We run one KS test for each of the 120 (i.e., $40 \times 3$) tasks, and perform p-value correction independently. Each of these statistical tests is performed on 200 demographic profiles in our experiment.

For both tests, we set 10 as the support threshold, aggregating conditions that appear fewer than 10 times into an ``\textit{others}'' class. Each of our tasks comprises 40 items (e.g., 40 essays for \textit{AES}).
For brevity, we focus on \textit{AES} to illustrate our setup. %the setup is the same for the three educational tasks, here we focus on AES as an example.
Using 200 user profiles from the \textit{AI Gap} dataset, we prompt the LLMs to perform each task conditioned on these demographic attributes. For instance, for the \implicit{} condition, each LLM scores each essay 200 times -- once for each implicit profile.
Because each user profile appears 40 times (once per essay), the data entries are not independent. 
To avoid violating the assumption of independence, we cannot run a single statistical test across the 8,000 data samples (40 items $\times$ 200 profiles). Instead, we conduct 40 separate statistical tests -- one per item -- and evaluate the significance of demographic attributes independently before aggregating the results.
An attribute is considered globally significant only if it achieves significance in at least $N$ individual tests. 
We set $N=10$ for \explicit{} and $N=5$ for \implicit{}; a lower threshold is used for \implicit{} because models must detect much subtler, less visible demographic differences.\footnote{The choice of $N$ for \explicit{} and \implicit{} is heuristic and is primarily intended to facilitate the presentation of the results.}
Additionally, in all tables in the results section (\S{}\ref{sec:results}), we report only effects for which the absolute average regression coefficient ($|\mu_b|$) exceeds 10\% of the standard deviation of the metric under consideration.

\subsection{Models}

We evaluate six instruction-tuned LLMs spanning proprietary and open-source families of varying sizes: GPT-5-mini, GPT-5-nano, Llama 3.3 70B, Llama 3.2 3B, Qwen3 4B, and Qwen3 30B. Exact model versions, hyperparameters, and references are provided in Appendix \ref{app:exp_setup}.
Each model processes every task under all three prompting conditions (\model{}, \implicit{}, and \explicit{}), enabling controlled within-model comparisons. The full experimental design comprises 6 models and 2 datasets (each with 40 items). 
For each dataset, every model is prompted with $(40 \times 200) \times 2$ conditioned prompts (for \implicit{} and \explicit{}), and with 40 baseline prompts (for \model{}). 
Overall, this results in $6 \times 2 \times [(40 \times 200) \times 2 + 40 ] = 192{,}480$ LLM inference calls. Approximately one-third of these calls correspond to API-based inference for proprietary models, while the remaining calls are executed on locally deployed open-weight models. %\footnote{Hyperparameters and implementation details are provided in Appendix~\ref{app:exp_setup}.} 
 In addition, \implicit{} requires a one-time simulation of 200 profile-specific interaction histories, which we generate using GPT-5-nano, resulting in 200 additional API calls; we use the same simulated responses for all models in our experiments.

\section{Results and Discussion}\label{sec:results}

\subsection{Automated Essay Scoring}
First, we compare the model's score for a given essay under \explicit{} and \implicit{} demographic conditions, and then contrast these scores with the \model{} outputs and the human baseline. \Cref{tab:score-deltas} shows the mean score deviation per model and condition, where the human mean score is $3.02 \pm 1.00$. Positive values indicate that the model assigns higher scores than humans under that condition. The Llama-3B significantly overscores relative to humans in the \model{} condition ($+0.48$, $p<0.01$), and Llama-70B underscores ($-0.52$, $p<0.01$). Under the \explicit{} condition, deviations from human ratings remain modest and directionally positive for most models, with significant upward shifts for Llama-3B ($+0.55$, $p<0.01$) and Qwen-4B ($+0.29$, $p<0.05$), and a significant downward shift for Llama-70B ($-0.60$, $p<0.001$). The \implicit{} condition reveals the starkest effect: Llama-70B scores inflate by $+1.57$ points relative to its own default ($p<0.001$), the largest deviation observed across all models and conditions. In addition, GPT models are the most stable, and neither \explicit{} nor \implicit{} personas produce significant shifts in scores relative to human ratings. 

\begin{table}[t]
\centering
\scriptsize
\setlength{\tabcolsep}{2pt}
\newcolumntype{Y}{>{\centering\arraybackslash}X}
\begin{tabularx}{\columnwidth}{lYYYYY}
\toprule
& \multicolumn{3}{c}{w.r.t.\ human} 
& \multicolumn{2}{c}{w.r.t.\ default} \\
\cmidrule(lr){2-4} \cmidrule(lr){5-6}
\textbf{Model} & \textbf{\model{}} & \textbf{\explicit{}} & \textbf{\implicit{}} & \textbf{\explicit{}} & \textbf{\implicit{}} \\
\midrule
GPT-mini     & $+0.05$ & $+0.16$ & $-0.02$ & $+0.11$ & $-0.07$ \\
GPT-nano     & $+0.10$ & $+0.25$ & $+0.10$ & $+0.15^{*}$ & $0.00$ \\
Llama-70B    & $-0.52^{**}$ & $-0.60^{***}$ & $+1.04^{***}$ & $-0.07$ & $+1.57^{***}$ \\
Llama-3B     & $+0.48^{**}$ & $+0.55^{**}$ & $-0.01$ & $+0.08$ & $-0.49^{***}$ \\
Qwen-30B     & $-0.05$ & $+0.11$ & $-0.15$ & $+0.16^{**}$ & $-0.10$ \\
Qwen-4B      & $+0.12$ & $+0.29^{*}$ & $-0.05$ & $+0.16^{**}$ & $-0.18^{**}$ \\
\bottomrule
\end{tabularx}
\caption{Mean AES score deviation from human ratings and \model{}, per condition. Significance assessed via paired $t$-test on 40 per-essay means $^{*}p{<}0.05$, $^{**}p{<}0.01$, $^{***}p{<}0.001$}
 \label{tab:score-deltas}
 \vspace{-15pt}
\end{table}
\begin{table}[ht]
\centering
\scriptsize
\begin{tabularx}{\columnwidth}{l X c r}
\toprule
\textbf{Model} & \textbf{Attribute} & $\mu{}_b\;(\sigma{}_b)$ & \textbf{Cnt} \\
\midrule
  \textbf{Qwen-30B} & \textbf{education} & $\mathbf{-0.11}$ $\mathbf{(0.08)}$ & \textbf{12} \\
  Qwen-4B & frequency\_llm & $+0.04$ $(0.07)$ & 15 \\
  % Qwen-30B & mum\_education & $-0.03$ $(0.02)$ & 11 \\
  % Qwen-4B & education & $-0.02$ $(0.09)$ & 14 \\
  % Qwen-4B & age & $-0.01$ $(0.06)$ & 11 \\
  % Qwen-4B & dad\_education & $-0.01$ $(0.03)$ & 11 \\
  % GPT-mini & ses & $+0.00$ $(0.04)$ & 12 \\
  % Llama-3B & education & $+0.00$ $(0.11)$ & 13 \\
  % Llama-3B & frequency\_llm & $+0.00$ $(0.03)$ & 12 \\
\midrule
  % \textbf{Llama-70B} & \textbf{know\_nlp: Text-to-Speech } & $\mathbf{+0.27}$ $\mathbf{(0.10)}$ & \textbf{5} \\
  \textbf{Llama-70B} & \textbf{usecases: Brainstorming} & $\mathbf{+0.26}$ $\mathbf{(0.16)}$ & \textbf{24} \\
  \textbf{Llama-70B} & \textbf{usecases: Writing} & $\mathbf{+0.24}$ $\mathbf{(0.08)}$ & \textbf{19} \\
  % \textbf{Llama-70B} & \textbf{tech: Laptop} & $\mathbf{+0.21}$ $\mathbf{(0.05)}$ & \textbf{5} \\
  \textbf{Llama-70B} & \textbf{religion: Nothing} & $\mathbf{+0.19}$ $\mathbf{(0.12)}$ & \textbf{20} \\
  \textbf{Llama-70B} & \textbf{hobbies: arts} & $\mathbf{+0.18}$ $\mathbf{(0.13)}$ & \textbf{15} \\
  \textbf{Llama-70B} & \textbf{would\_nlp: Dialog tech.} & $\mathbf{+0.17}$ $\mathbf{(0.12)}$ & \textbf{20} \\
  % \textbf{Llama-70B} & \textbf{hobbies: museums} & $\mathbf{+0.16}$ $\mathbf{(0.11)}$ & \textbf{6} \\
  % \textbf{Llama-70B} & \textbf{mum\_occupation: Service and sales workers} & $\mathbf{+0.16}$ $\mathbf{(0.07)}$ & \textbf{8} \\
  % \textbf{Llama-70B} & \textbf{llm\_use: GitHub Copilot} & $\mathbf{+0.12}$ $\mathbf{(0.08)}$ & \textbf{7} \\
  \textbf{Llama-70B} & \textbf{gender: Female} & $\mathbf{+0.12}$ $\mathbf{(0.08)}$ & \textbf{21} \\
\bottomrule
\end{tabularx}
\caption{Impact of different demographic attributes in \explicit{} (top) and \implicit{} (bottom) on the score of the model responses, ordered by $|\mu_b|$. In bold if $|\mu_b| \geq \sigma_\text{resp.score}$}
\label{tab:results_regression_test_aes_short}
\end{table}

Our regression analysis shows multiple significant results for scoring in the AES task (summary of attributes with $\textbf{Cnt}\ge10$ in  \Cref{tab:results_regression_test_aes_short}\footnote{We note that attributes like ``language:English'' in the \textit{AI Gap} dataset are majority-group attributes and should be interpreted as deviations from a demographically skewed baseline, not as genuine group contrasts.}). Across all models in the \explicit{} condition, demographic attributes generally produce only small shifts in scores. The largest observed effect is in \explicit{}, Qwen-30B penalizes scores from users with higher education levels ($\mu_b = -0.11$). In \implicit{}, Llama-70B constantly overscores essays based on different demographic attributes. Users mentioning brainstorming ($+0.26$) or writing ($+0.24$) in their conversation history receive notably higher scores\footnote{The complete regression results of the AES task are available in \Cref{tab:results_regression_test_aes_all}}. 
The KS test compares score distributions across demographic subgroups and assesses whether demographic attributes shift the full score distribution — not merely its mean. We find significant results only in \explicit{} condition. For instance, for education Llama-3B: $D(\text{edu}_2,\text{edu}_0) = 0.65$ and Qwen-30B: $D(\text{edu}_2,\text{edu}_0) = 0.60$ produce large distributional differences, with effects tending to grow across more distant ordinal levels (details in \Cref{tab:ks-aes}).

\subsection{Metalinguistic QA}\label{subsec:qa_results}
As described in \S{}\ref{subsec:evaluation}, for \textit{QA} we use a variety of metrics to measure demographic sensitivity. Unlike the AES task, where LLM responses are anchored to a fixed evaluation rubric, the QA task is inherently less constrained. There is no single objectively correct response, leaving the model greater latitude to adapt its tone, complexity, and framing to perceived user characteristics. We therefore anticipate demographic-driven variation in QA response metrics to be larger than in AES (\textit{RQ2}).

\paragraph{Text accessibility}\label{subsubsec:readability}
We found multiple significant results for the two readability indexes, for both \explicit{} (more frequently) and \implicit{}.
\begin{table}[ht]
\centering
\scriptsize
\begin{tabular}{llcr}
\toprule
 % &  & & Adj.\\
\textbf{Model} & \textbf{Attribute} & $\mu{}_b (\sigma{}_b)$ & \textbf{Cnt}\\
% ARI
\midrule
Llama-70B & education & $+0.73$ $(0.51)$ & 36 \\
Llama-70B & nationality: US & $+0.33$ $(0.39)$ & 20 \\
Llama-70B & language: English & $+0.31$ $(0.37)$ & 12 \\
Llama-70B & nationality: UK & $-0.29$ $(0.52)$ & 13 \\
Llama-3B & empl: Employed full time & $+0.29$ $(0.26)$ & 11 \\
Llama-3B & age & $-0.28$ $(0.20)$ & 16 \\
Llama-3B & education & $+0.25$ $(0.19)$ & 14 \\
Llama-70B & empl: Employed full time & $+0.24$ $(0.20)$ & 22 \\
Llama-70B & mum\_education & $+0.24$ $(0.40)$ & 21 \\
Qwen-30B & education & $+0.21$ $(0.12)$ & 20 \\
Llama-3B & nationality: US & $+0.19$ $(0.30)$ & 12 \\
Llama-70B & frequency\_llm & $-0.17$ $(0.29)$ & 10 \\
GPT-mini & education & $+0.16$ $(0.12)$ & 13 \\
Llama-70B & ses & $+0.16$ $(0.10)$ & 18 \\
Qwen-30B & empl: Employed full time & $+0.12$ $(0.10)$ & 13 \\
Llama-70B & dad\_education & $+0.11$ $(0.06)$ & 16 \\
\midrule
Llama-3B & gender: Male & $+0.53$ $(0.23)$ & 5 \\
Qwen-4B & religion: Christian & $+0.31$ $(0.24)$ & 5 \\
GPT-mini & language: English & $+0.21$ $(0.16)$ & 6 \\
Llama-70B & language: English & $+0.20$ $(0.12)$ & 6 \\
GPT-nano & gender: Male & $+0.19$ $(0.17)$ & 5 \\
Qwen-30B & language: English & $+0.19$ $(0.11)$ & 6 \\
Qwen-4B & language: English & $+0.16$ $(0.11)$ & 5 \\
Qwen-4B & gender: Male & $+0.12$ $(0.07)$ & 5 \\
\bottomrule
\end{tabular}
\caption{Impact of different demographic attributes in \explicit{} (top) and \implicit{} (bottom) on the ARI of the model responses, ordered by $|\mu_b|$. 
% $b$ is the coefficient obtained from the regression test, we show its mean ($\mu$) and standard deviation ($\sigma$) across the tasks for which it proved significant. 
% We show only significant results, with $|\mu_b| \geq 0.1 \times\sigma{}_\text{ARI}$, and support $\geq 10$ (explicit) or $\geq 5$ (implicit).
% Larger ARI indicates less readable texts. % (range is $[0; 15$); larger FRE indicates more readable texts (range is $[0; 100$). 
% (in bold $\geq 1 \times \sigma{}$ the standard deviation).
}
\label{tab:results_qa_readability_regression_test_ari}
\end{table}

\Cref{tab:results_qa_readability_regression_test_ari} shows how the \textbf{ARI} (larger values indicates more difficult texts) of the model response changes for different demographics, without considering \model{}. Notably, for categorical attributes, the coefficient indicates how much the readability index changes for the prompts conditioned with a demographic attribute with respect to the other attributes; for the ordinal attributes (e.g. \textit{education}) the coefficient indicates how much the value of the readability index changes moving to the following value in the list (e.g., from \textit{High school or below} to \textit{undergraduate}).
The responses provided by the GPT and Qwen models rarely change significantly, and when they do, the coefficient is generally smaller than that of the two Llama models. On the other hand, the readability of the responses of two Llama models changes significantly for a variety of attributes -- although the coefficient is consistently smaller than the standard deviation of the ARI ($\sigma_\text{ARI} = 1.06$).
The \textit{Education} attribute consistently leads to significant positive coefficients for \explicit{} (for Llama-70B, Llama-3B, Qwen-30B, and GPT-mini), meaning that users with higher educational level receive less readable responses: this suggests that if the models have access to an explicit reference to the educational level of the user they can partially adapt the readability of their responses.
A similar effect is visible, to a lesser extent and for Llama-70B only, for \textit{mum\_education}, \textit{ses}, and \textit{dad\_education}. 
However, this adaptation does nor occur for \implicit{}: indeed, we find that users whose first language is English receives (slightly) less readable responses (from GPT-mini, Llama-70B, and the two Qwen models), as well as users who identified as Males (from Llama-3B, GPT-nano, and Qwen-4B), and Christians (Qwen-4B). We note that this might be due to the effect of the majority-group attributes. Focusing on the results of the KS test (details in \S{}\ref{sec:app:qa:ari}) we find that, for the same model, the difference in readability tends to be larger for larger \textit{ses} gaps (Llama-70B: $D(\text{ses}_4,\text{ses}_8) = 0.79$, $D(\text{ses}_4,\text{ses}_7) = 0.63$) and for larger \textit{education} gaps (e.g., for Qwen-30B: $D(\text{edu}_0,\text{edu}_2) = 0.51$, $D(\text{edu}_0,\text{edu}_1) = 0.42$, $D(\text{edu}_1,\text{edu}_2) = 0.39$), although this is not uniform across levels.

% FRE
The results for \textbf{FRE} are similar to those for ARI (full details and tables in \S{}\ref{sec:app:qa:fre}).
According to the regression test (\Cref{tab:results_qa_readability_regression_test_fre}), higher education levels consistently receive responses of lower readability for \explicit{} (we found significance for 5 models), and we also find the same effect with \implicit{} on Qwen-4B (but with a smaller coefficient).
Similarly, users with higher \textit{SES} values receive less readable responses in both \explicit{} (4 out of 6 models) and \implicit{} (3 models). As for the ARI, the KS test shows that larger gaps in education levels lead to larger differences in the readability of the LLM response (Table \ref{tab:results_qa_readability_ks_test_fre}).
Comparing the effect of the same attributes in the two conditioning settings, we find that the demographic attributes tend to have the same effect, but with smaller magnitude for \implicit{}: indeed, the coefficient of the least-squares regression line (\explicit{} on the $x$-axis, more details and plots in \S{}\ref{sec:app:metalinguistic_qa}) is $\mu_b = 0.56$ for ARI and $\mu_b = 0.51$ for FRE.

% Response length
\begin{table}[h]
\centering
\scriptsize
\begin{tabular}{llcr}
\toprule
 % &  & & Adj.\\
\textbf{Model} & \textbf{Attribute} & $\mu{}_b (\sigma{}_b)$ & \textbf{Cnt}\\
\midrule
Llama-70B & language: English & $+7.83$ $(4.08)$ & 20 \\
Qwen-30B & education & $+7.11$ $(10.90)$ & 12 \\
Llama-70B & education & $+6.89$ $(6.28)$ & 29 \\
Llama-70B & nationality: US & $+6.41$ $(7.11)$ & 19 \\
Llama-3B & education & $+4.25$ $(2.99)$ & 11 \\
GPT-mini & education & $+4.22$ $(4.22)$ & 11 \\
Qwen-30B & mum\_education & $+4.19$ $(3.67)$ & 10 \\
Llama-70B & gender: Male & $+3.50$ $(2.71)$ & 10 \\
Llama-70B & mum\_education & $+3.43$ $(2.67)$ & 23 \\
\midrule
\textbf{Qwen-4B} & \textbf{tech: Laptop} & $+32.97$ $(27.88)$ & \textbf{5} \\
\textbf{Qwen-30B} & \textbf{usecases: Writing} & $+30.64$ $(34.21)$ & \textbf{5} \\
GPT-mini & tech: Laptop & $+21.20$ $(21.87)$ & 5 \\
Llama-3B & tech: Laptop & $+13.94$ $(14.12)$ & 6 \\
GPT-nano & tech: Laptop & $+12.08$ $(11.88)$ & 5 \\
Llama-70B & hobbies: Exercise & $+9.65$ $(9.93)$ & 7 \\
GPT-nano & tech: Tablet & $+9.57$ $(10.71)$ & 5 \\
GPT-mini & gender: Male & $+8.49$ $(12.33)$ & 5 \\
GPT-nano & empl: Employed full time & $+8.49$ $(5.19)$ & 5 \\
GPT-mini & empl: Employed full time & $+8.41$ $(4.58)$ & 6 \\
GPT-nano & hobbies: Use social media & $+7.21$ $(6.26)$ & 5 \\
Qwen-4B & usecases: Coding & $-6.38$ $(7.34)$ & 5 \\
Llama-70B & language: English & $+5.51$ $(5.23)$ & 8 \\
GPT-mini & mum\_education & $+3.18$ $(2.98)$ & 5 \\
\bottomrule
\end{tabular}
\caption{Impact of different demographic attributes in \explicit{} (top) and \implicit{} (bottom) on the length (number of words) of the model responses, ordered by $|\mu_b|$. In bold if $|\mu_b| \geq \sigma_\text{resp.length}$.
% Impact of different explicit (top) and implicit (bottom) conditions on the length, measured as the number of words, of the model responses, ordered by magnitude. 
% $b$ is the coefficient obtained from the regression test, we show its mean ($\mu$) and standard deviation ($\sigma$) across the tasks for which it proved significant. 
% We show only coefficients with $|\mu_b| \geq 0.1 \times\sigma{}_\text{resp.length}$ (in bold when $|\mu_b| \geq \sigma_\text{resp.length}$), and support $\geq 10$ (explicit) or $\geq 5$ (implicit).
}
\vspace{-15pt}
\label{tab:results_qa_response_length_ols_test}
\end{table}

Notably, our findings were quite different when measuring differences in the \textbf{response length} (number of words) of the LLM responses: indeed, we find  more significant results for \implicit{} than for \explicit{}, and the coefficients were generally larger, as shown in Table \ref{tab:results_qa_response_length_ols_test}.
We consistently find that users who use \textit{laptops} receive longer responses (4 models).
% , which might suggests that the models can adapt the length of their responses to the length of the previous histories of the users. 
Again, we find that the responses for higher education levels are longer (which is potentially correlated with less readable texts), and we observe the same effect (with smaller coefficients) for \textit{mum\_education} in Qwen-30B and Llama-70B.
The fact that \implicit{} has stronger effects is also visible from the least-squares regression line (\Cref{fig:resp_length:imp_vs_exp} in \S{}\ref{sec:app:qa:resplength}), which has a coefficient $m=1.20$.
Conversely, we did not find many attributes who consistently appeared as significant according to the KS test; indeed, we only found \textit{education} for Llama-70B when conditioning explicitly:  $D(\text{edu}_0,\text{ses}_2) = 0.54 (\pm0.09)$ and  $D(\text{edu}_0,\text{edu}_1) = 0.46 (\pm0.06)$, with a support of 13 and 8 tasks respectively. Lastly, our statistical tests yield few significant results for \textbf{AoA}, and we report the statistically significant attributes in \S{}\ref{sec:app:qa:aoa}.

% Conclusion for the readability analysis
Overall, our readability analysis suggests that when \explicit{} references to the user's (or their parents') education level or Socio-Economic Status are available, most models partially adapt their responses, producing less readable messages for users with higher education levels and SES.
However, the same adaptation is not present in \implicit{} conditioning, and we only find that the models often mimic surface characteristics of previous prompt histories (e.g., longer responses for users with longer prompts in their history).

% Sentiment analysis
\paragraph{Sentiment Analysis.}
\begin{table}[ht]
\vspace{-10pt}
\centering
\scriptsize
\begin{tabular}{llcr}
\toprule
 % &  & & Adj.\\
\textbf{Model} & \textbf{Attribute} & $\mu{}_b (\sigma{}_b)$ & \textbf{Cnt}\\ 
\midrule
\textbf{Llama-70B} & \textbf{education} & $+0.08$ $(0.06)$ & \textbf{26} \\
\textbf{Llama-3B} & \textbf{education} & $+0.07$ $(0.09)$ & \textbf{12} \\
Llama-70B & empl: Employed full time & $+0.05$ $(0.02)$ & 12 \\
Llama-3B & age & $+0.04$ $(0.07)$ & 14 \\
Llama-70B & ses & $+0.02$ $(0.03)$ & 12 \\
\midrule
\textbf{Llama-3B} & \textbf{tech: Smartphone} & $+0.19$ $(0.19)$ & \textbf{5} \\
\textbf{Qwen-30B} & \textbf{religion: Christian} & $+0.08$ $(0.13)$ & \textbf{5} \\
\textbf{GPT-nano} & \textbf{hobbies: Use social media} & $+0.07$ $(0.05)$ & \textbf{5} \\
Qwen-4B & nationality: US & $+0.05$ $(0.03)$ & 5 \\
GPT-mini & empl: Employed full time & $+0.03$ $(0.02)$ & 5 \\
Qwen-30B & education & $+0.02$ $(0.02)$ & 5 \\
Llama-70B & frequency\_llm & $-0.02$ $(0.02)$ & 5 \\
Llama-3B & ses & $+0.01$ $(0.03)$ & 6 \\
\bottomrule
\end{tabular}
\caption{Impact of different demographic attributes in \explicit{} (top) and \implicit{} (bottom) on the sentiment of the model responses, ordered by $|\mu_b|$. In bold if $|\mu_b| \geq \sigma_\text{sentiment}$.
% Impact of different explicit (top) and implicit (bottom) conditions on the sentiment of the model responses, ordered by magnitude. 
% $b$ is the coefficient obtained from the regression test, we show its mean ($\mu$) and standard deviation ($\sigma$) across the tasks for which it proved significant. 
% We show only coefficients with $|\mu_b| \geq 0.1 \times\sigma{}_\text{sentiment}$ (in bold when $|\mu_b| \geq \sigma_\text{sentiment}$), and support $\geq 10$ (explicit) or $\geq 5$ (implicit).
}
\vspace{-5pt}
\label{tab:results_qa_response_sentiment_regression_test}
\end{table}

Table \ref{tab:results_qa_response_sentiment_regression_test} presents the results of the multivariate regression test on the sentiment of the conditioned model responses. We found significance for both \implicit{} and \explicit{}, and in some cases for the same conditions: \textit{education} leads to a positive coefficient for the two Llama models (\explicit{}) and, with smaller coefficient, Qwen-30B (\implicit{}). This indicates that higher education level receives responses with more positive sentiment, which can be problematic especially with \explicit{}: indeed, the dataset contains 4 education levels (from \textit{`High school or below'} to \textit{`Doctorate or above'}) and the coefficient indicates that, on average, there is a difference of $0.3$ between the sentiment of the lowest and higher education level (with $\text{sentiment} \in \{0,1,2,3,4\}$, and $\sigma_\text{sentiment} = 0.07$). 
The same effect is visible for \implicit{} with Qwen-30B, but with a smaller coefficient.
We also find that \textit{`Employment: Employed full time'} leads to positive coefficients in both \explicit{} (Llama-70B) and \implicit{} (GPT-mini), as well as \textit{ses} (Llama-70B explicit and Llama-3B implicit), both with a coefficient which is smaller for \implicit{}.
Finally, we find that for \implicit{} there are several attributes which have a strong effect on the sentiment of the responses ($|\mu_b| \geq \sigma_\text{sentiment}$) even though they do not appear from the analysis on \explicit{}: \textit{Tech: Smartphone}, \textit{religion: Christian}, \textit{hobbies: Use social media}.
Most likely, LLMs pick up the topics in these users' previous prompt histories, leading to more positive sentiments; however, the effects are not consistent across models.

\paragraph{Topic Analysis.}
To further investigate potential sources of variation in the implicit demographic condition, we analyze the topic distribution of users' prompt histories. Among the 15 identified topics, the most prevalent are \textit{General Q\&A}, \textit{Career \& Resume}, and \textit{AI \& Concept Explanation}. The complete topic-modeling results are presented in \Cref{fig:topic-model}. The demographic breakdown of topic distributions reveals systematic differences in the types of queries. Gender differences are apparent: women's prompts are more concentrated in \textit{Career \& Resume}, whereas men's prompts are more frequently associated with \textit{General Q\&A}. Age-related differences are also observed, with younger users (18--34) showing a greater concentration of AI-related queries (T11, 12.8\%), while older users (45--54 and 60+) are more concentrated on \textit{Account \& Admin} tasks (T7, up to 15.5\%). Socioeconomic status exhibits a similar pattern, with lower-SES users more frequently associated with \textit{Factual Lookup} (T4, 24.0\% at SES 1), whereas higher-SES users show a greater concentration of \textit{Finance \& Stocks} queries (T2, 23.3\% at SES 10). Similar differences are observed across employment status and frequency of AI usage.

\paragraph{BERTScore.}
Differently from the metrics discussed previously, which only focus on surface aspects (e.g., readability, length, ...), BERTScores can capture the semantics of the model responses.
\begin{table}[ht]
\centering
\scriptsize
\begin{tabular}{llcr}
\toprule
 % &  & & Adj.\\
\textbf{Model} & \textbf{Attribute} & $\mu{}_b (\sigma{}_b)$ & \textbf{Cnt}\\ 
\midrule
\textbf{GPT-mini} & \textbf{language: English} & $+0.0150$ $(0.0090)$ & \textbf{14} \\
\textbf{Llama-70B} & \textbf{language: English} & $+0.0116$ $(0.0060)$ & \textbf{31} \\
Llama-70B & education & $+0.0069$ $(0.0039)$ & 36 \\
Llama-70B & nationality: US & $+0.0040$ $(0.0052)$ & 23 \\
Qwen-4B & language: English & $+0.0038$ $(0.0040)$ & 16 \\
Llama-3B & education & $+0.0036$ $(0.0040)$ & 12 \\
Llama-3B & age & $-0.0033$ $(0.0023)$ & 20 \\
GPT-nano & language: English & $+0.0032$ $(0.0030)$ & 10 \\
Llama-70B & age & $-0.0032$ $(0.0020)$ & 27 \\
GPT-mini & nationality: US & $+0.0032$ $(0.0019)$ & 11 \\
Llama-3B & nationality: US & $+0.0029$ $(0.0050)$ & 15 \\
Llama-3B & dad\_education & $+0.0018$ $(0.0017)$ & 15 \\
Llama-70B & dad\_education & $+0.0018$ $(0.0015)$ & 10 \\
Llama-70B & mum\_education & $+0.0016$ $(0.0012)$ & 20 \\
Qwen-4B & age & $-0.0014$ $(0.0021)$ & 17 \\
Llama-70B & ses & $+0.0014$ $(0.0013)$ & 12 \\
Qwen-30B & nationality: US & $+0.0014$ $(0.0020)$ & 10 \\
\midrule
Llama-70B & hobbies: social media & $+0.0063$ $(0.0090)$ & 5 \\ % was: use social media
Llama-3B & language: English & $+0.0028$ $(0.0014)$ & 5 \\
GPT-mini & tech: Laptop & $+0.0026$ $(0.0012)$ & 6 \\
Llama-70B & language: English & $+0.0022$ $(0.0016)$ & 7 \\
Qwen-4B & gender: Male & $+0.0020$ $(0.0008)$ & 5 \\
Llama-70B & ses & $-0.0019$ $(0.0010)$ & 5 \\
GPT-nano & use\_nlp: detect spam & $+0.0016$ $(0.0021)$ & 5 \\  % was: use\_nlp: E-mail spam detection
Qwen-30B & language: English & $+0.0016$ $(0.0009)$ & 8 \\
Llama-70B & age & $+0.0012$ $(0.0010)$ & 6 \\
GPT-nano & home: Rent & $+0.0010$ $(0.0010)$ & 5 \\
Qwen-4B & age & $+0.0010$ $(0.0005)$ & 8 \\
\bottomrule
\end{tabular}
\caption{Impact of different demographic attributes in \explicit{} (top) and \implicit{} (bottom) on the BERTscore F1 of the model responses (using \model{} as reference), ordered by $|\mu_b|$. In bold if $|\mu_b| \geq \sigma_\text{BERTscoreF1}$.
% Impact of different explicit (top) and implicit (bottom) conditions on the BERTscore F1 of the model responses (using the \textit{model default} as reference), ordered by magnitude. 
% $b$ is the coefficient obtained from the regression test, we show its mean ($\mu$) and standard deviation ($\sigma$) across the tasks for which it proved significant. 
% We show only coefficients with $|\mu_b| \geq 0.1 \times\sigma{}_\text{BERTscoreF1}$ (in bold when $|\mu_b| \geq \sigma_\text{BERTscoreF1}$), and support $\geq 10$ (explicit) or $\geq 5$ (implicit).
}
\label{tab:results_qa_bertscore_f1_regression_test}
\end{table}

\Cref{tab:results_qa_bertscore_f1_regression_test} shows the results of the regression test on the BERTscore F1 of the LLM responses, using \model{} as reference (i.e., we measure how much the response of the model is steered away from the default under the different prompting conditions; larger BERTscore values indicate responses that are more similar to the reference).
We find that, beyond \textit{language:English} (the majority attribute), \textit{nationality: US}, shows positive coefficients for \explicit{} conditioning (4 models). These results show, once more, that the \textit{default} of LLMs tends to be English-centric (and US-centric in particular). 
We also find that higher \textit{education} levels lead to responses more similar to the default when stated explicitly, as well as lower \textit{age} and higher \textit{SES}.
Notably, we find a contrasting effect for the latter two: their effect is opposite with \implicit{} and \explicit{}: \textit{age} has a negative coefficient for \explicit{} (3 models) but positive for \implicit{} (2 models), while \textit{ses} has a positive coefficient for \explicit{} but a negative for \implicit{} (same model, Llama-70B, in both cases).

\subsection{Formative 
Feedback}\label{subsec:ff_results}
Similar to \S{}\ref{subsec:qa_results}, we report the most significant attributes from regression analysis for the FF task.\footnote{Details of all metrics and the KS-test are in \S{}\ref{sec:app:aes}} 

\paragraph{Text accessibility.} In essay feedback generation, all significant attributes are driven almost entirely by \implicit{} conditioning and are largely concentrated in Llama-70B. \textbf{FRE.} Personas associated with \textit{usecases: Writing} ($\mu_b{=}{+}3.67$, Cnt$=$6), \textit{hobbies: museums or exercise} ($+$1.93 to $+$1.72), \textit{religion:Christian} ($+$1.41), and \textit{gender: Female} ($+$1.09) all receive more readable feedback with no \explicit{} effect. For \textbf{ARI} (\Cref{tab:ff-ols-ari}), the \implicit{} effects are sparse, and \textit{hobbies: museums or exercise} appears again for Llama-70B, GPT-nano.
 
\paragraph{Response length.} Under \implicit{} condition, Qwen-4B shows
longer feedback for behavioral attributes such as brainstorming use cases ($+$2.51), knowledge of dialog technology ($+$2.46), and digital content creation ($+$2.41). Notably, Qwen-4B also shows a significant effect for \textit{ethnicity: Black or African American} ($+$1.35, Cnt$=$5)---the only core demographic attribute beyond technology-use patterns to emerge in this metric.
 
\paragraph{Sentiment.} No \explicit{} effects are found (\Cref{tab:ff-ols-sentiment}). Under \implicit{} all attributable exclusively to Llama-70B. We observe significant over behavioral attributes such as (\textit{usecase:Brainstorming}: $+$0.32; \textit{tech:laptop}: $+$0.24; \textit{usecase:Writing}: $+$0.19)) and core demographics such \textit{(religion:Noting)}, (\textit{gender:Female}: $+$0.14). The breadth of attributes suggests that Llama-70B's feedback sentiment is broadly susceptible to implicit persona context.

\section{Related Work}
LLMs are increasingly being adopted for educational
tasks such as automated scoring, feedback generation, tutoring, question
generation, and learning analytics, offering opportunities for scalable and
personalized instructional support \citep{parker2025large,KASNECI2023102274,
nkoyoadvances,gupta-etal-2025-large,golchin-etal-2025-grading,stahl-etal-2024-exploring,10.1145/3698205.3729551}. However, their deployment in educational settings raises important concerns about validity, reliability, transparency, and fairness, particularly when model outputs influence assessment, feedback, or student learning trajectories \citep{LIU2026100565,gupta-etal-2025-large,gallegos-etal-2024-bias}. A central risk is that LLMs may reproduce or amplify demographic, linguistic, cultural, and socioeconomic biases inherited from training data in explicit and implicit conditions \citep{warr2025implicit,jin-etal-2024-implicit,weissburg2025llms,rooein-etal-2025-biased,zhao-etal-2025-explicit}. Most existing studies focus only on a limited set of explicit demographic attributes such as race, gender, and socioeconomic status \cite{du2025benchmarking,shailya-etal-2025-study,warr2025analyzing, du2025benchmarking,gandara2024inside}. In this study, however, we examine the effects of a broad range of explicit demographic attributes, together with implicit conversational characteristics derived from real user interactions with LLMs, across three educational tasks.

\section{Conclusion}
We systematically investigate how explicit and implicit demographic signals influence LLM behavior across three educational assessment tasks. Using a controlled experimental setup, we evaluate 6 LLMs, 25 demographic attributes, and 10 rounds of conversation across 200 real user profiles. Our findings reveal that demographic effects are strongly task-dependent. Most models remain relatively stable given \explicit{} persona conditioning in AES. However, \implicit{} demographic cues produce larger and less predictable effects, particularly for Llama-70B model. In contrast, for open-ended tasks, models show systematic differences in readability, sentiment, and response length. Our findings have important implications for using AI in performing educational tasks such as scoring and generating feedback for students: in high-stakes evaluative settings, demographic invariance is essential. However,  in instructional contexts, some adaptation may be pedagogically appropriate, such as adjusting readability metrics. However, current LLMs cannot reliably separate task-relevant information from demographic cues. That raises concerns about fairness, transparency, and the potential amplification of educational inequities. As LLMs become increasingly integrated in education, we need to ensure their behavior reflects pedagogical objectives rather than unintended demographic biases.

\section{Acknowledgments}
This work was conducted while Donya Rooein was a Postdoctoral Researcher at Bocconi University. Donya Rooein and Dirk Hovy were supported by the European Research Council (ERC) under the European Union’s Horizon 2020 research and innovation program (grant agreement No. 949944, INTEGRATOR) and MUR FARE 2020 initiative under grant agreement Prot. R20YSMBZ8S (INDOMITA).

\section*{Limitations}
This work has some limitations in terms of the dataset used for the evaluation: indeed, to limit the computational costs of the experiments, we subsampled the original datasets and kept only 120 items (40 each from Automated Essay Scoring, Formative Feedback, and Metalinguistic QA), and 200 user profiles (subsampled from the \textit{AI Gap} paper).
Although the statistical tests we consider account for this when studying the significance, this work would benefit from larger datasets, and potentially from the inclusion of other educational tasks. In addition, several demographic attributes in the AI Gap dataset are heavily skewed toward majority groups (e.g., 96\% of profiles report English as their first language), making the corresponding regression coefficients uninterpretable as genuine group contrasts; we therefore exclude these from our results.

We use the prompt histories available in the original paper from \newcite{bassignana2025ai} for implicit conditioning; we do this in order to work with realistic prompts, but the histories are not necessarily related to the educational domain, and thus not necessarily representative of how LLMs are used by students in educational settings.
Indeed, we work on a dataset that might contain a lot of noise and thus make our study (of the effect of the implicit conditioning) even more challenging.
Another point to note is the fact that the tasks we prompt the models to perform (e.g., the essays which have to be scored) have been written by users whose demographic profile is different from those we give to the models with conditioning (both implicit and explicit conditioning).
This might lead to a conflicting signal: on one hand, there is the linguistic signal from the student essay, and, on the other, there is the linguistic signal from the history of previous prompts (or the explicit persona), and the two might ``pull'' in different directions. However, this does not affect the validity of our findings as the (potential) signal from the educational task is the same when conditioning on different demographics.
Also, even though we consider implicit conditioning, the fact that we experiment on English only implies that we cannot properly evaluate the effect of grammatical gender, which is much more significant in other languages \cite{benedetto2026beyond,ducel2025you}.

We use Bonferroni correction to adjust the p-value of the KS test, which has been criticized as being \textit{too conservative} \cite{nakagawa2004farewell} and thus missing some significant results. While this is arguably a limitation, since there might be other effects which have gone unnoticed, it suggests that the statistically significant results found in this paper are due to a real effect of the conditions on the model responses.

\section*{Ethics Statements}
This research examines demographic sensitivity in LLMs applied to educational tasks—a domain with significant equity implications. While our study \textbf{does not directly involve human subjects}, we use real user demographic profiles from the publicly available AI Gap dataset \citep{bassignana2025ai}, which was collected under appropriate ethical oversight and with informed consent. Our experimental design deliberately measures demographic sensitivity in LLMs for performing three educational tasks to identify potential harms before deployment. 
\bibliography{latex/custom}

\appendix

\section{List of demographic attributes}\label{sec:app:list_conditions}
The demographic attributes considered in this paper are a combination of single-value categorical, multi-value categorical, and ordinal attributes.

The ordinal attributes are \textit{age, ses, education, mum\_education, dad\_education, frequency\_llm}; they are provided as strings in the original \textit{AI Gap} dataset, but we convert them to integers (from $0$ to $N$) due to their intrinsic ordinal nature (e.g., \textit{Doctorate or above} is a higher educational level with respect to \textit{Undergraduate degree}).
They are listed in Table \ref{tab:appendix_ordinal_values}, together with their possible values.
In some cases, the value of one or more of these attributes is \textit{Prefer not to say}, which we replace with the median of the other values for the statistical test.
\begin{table}[h]
\scriptsize
\setlength{\tabcolsep}{4pt}
\begin{center}
\begin{tabular}{llcr}
\toprule
\textbf{Attribute} & \textbf{Values} & \textbf{Value (int)} & \textbf{Count}\\
\midrule
age & 18-24 & 0 & 36 \\
age & 25-34 & 1 & 60 \\
age & 35-44 & 2 & 45 \\
age & 45-54 & 3 & 40 \\
age & 55-60 & 4 & 8 \\
age & 60+ & 5 & 11 \\
\midrule
education & High school or below & 0 & 57 \\
education & Undergraduate degree & 1 & 77 \\
education & Graduate degree & 2 & 56 \\
education & Doctorate or above & 3 & 8 \\
education & Prefer not to say & - & 2 \\
\midrule
mum\_education & High school or below & 0 & 108 \\
mum\_education & Undergraduate degree & 1 & 45 \\
mum\_education & Graduate degree & 2 & 37 \\
mum\_education & Doctorate or above & 3 & 8 \\
mum\_education & Prefer not to say & - & 2 \\
\midrule
dad\_education & High school or below & 0 & 103 \\
dad\_education & Undergraduate degree & 1 & 41 \\
dad\_education & Graduate degree & 2 & 34 \\
dad\_education & Doctorate or above & 3 & 17 \\
dad\_education & Prefer not to say & - & 5 \\
\midrule
ses & 1 & 0 & 5 \\
ses & 2 & 1 & 18 \\
ses & 3 & 2 & 36 \\
ses & 4 & 3 & 25 \\
ses & 5 & 4 & 30 \\
ses & 6 & 5 & 32 \\
ses & 7 & 6 & 27 \\
ses & 8 & 7 & 15 \\
ses & 9 & 8 & 8 \\
ses & 10 & 9 & 3 \\
ses & Prefer not to say & - & 1 \\
\midrule
frequency\_llm & Rarely & 0 & 23 \\
frequency\_llm & Sometimes & 1 & 49 \\
frequency\_llm & Nearly every day & 2 & 69 \\
frequency\_llm & Every day & 3 & 59 \\
\bottomrule
\end{tabular}
\caption{List of the ordinal demographic attributes an their possible values, sorted by increasing value after conversion to integer. We refer to the original \textit{AI Gap} paper \cite{bassignana2025ai} for more details.}
\label{tab:appendix_ordinal_values}
\end{center}
\end{table}

The single-value categorical attributes are \textit{gender, nationality, ethnicity, marital, language, religion, home, employment}; their possible values are listed in Table \ref{tab:appendix_single_value_categorical_values}. 
For \textit{nationality, ethnicity}, and \textit{language} we opted to create a \textit{multi\_*} attribute (i.e., \textit{multi\_nationality, multi\_ethnicity, multi\_language}) for participants who selected multiple options.
This was done to reduce the degrees of freedom for the statistical tests, and because most of those options appeared rarely in the dataset, thus would be discarded due to the low support during the statistical tests (we only keep attributes that appear in at least 10 different user profiles).
In any case, most of them are dropped from the statistical tests due to the low support even after grouping.
\begin{table}[h]
\scriptsize
\setlength{\tabcolsep}{4pt}
\begin{center}
\begin{tabular}{llr}
\toprule
\textbf{Attribute} & \textbf{Values} & \textbf{Count}\\
\midrule
gender & Male & 108 \\ % 
gender & Female & 91 \\ % 
% gender & Non-binary & 1 \\ % 
\midrule
nationality & United States & 104 \\ % 
nationality & United Kingdom & 73 \\ % 
nationality & multi\_nationality & 10 \\ % 
% nationality & Canada & 3 \\ % 
% nationality & United Kingdom;Nigeria & 3 \\ % 
% nationality & Nigeria & 3 \\ % 
% nationality & United States;Canada & 2 \\ % 
% nationality & United States;Costa Rica & 1 \\ % 
% nationality & United States;Ghana & 1 \\ % 
% nationality & United States;Poland & 1 \\ % 
% nationality & United Kingdom;Kenya & 1 \\ % 
% nationality & Hungary & 1 \\ % 
% nationality & Zimbabwe & 1 \\ % 
% nationality & Australia & 1 \\ % 
% nationality & Egypt & 1 \\ % 
% nationality & South Africa & 1 \\ % 
% nationality & Kenya & 1 \\ % 
% nationality & United States;Australia & 1 \\ % 
\midrule
ethnicity & White / Caucasian & 119 \\ % 
ethnicity & Black or African American & 53 \\ % 
ethnicity & multi\_ethnicity & 13 \\ % 
ethnicity & Asian / Pacific Islander & 8 \\ % 
ethnicity & Hispanic & 5 \\ % 
% ethnicity & White / Caucasian;Asian / Pacific Islander & 3 \\ % 
% ethnicity & White / Caucasian;Black or African American & 2 \\ % 
% ethnicity & Asian / Pacific Islander;White / Caucasian & 2 \\ % 
% ethnicity & Other. Please specify & 2 \\ % 
% ethnicity & Hispanic;White / Caucasian & 1 \\ % 
% ethnicity & Black or African American;Other. Please specify & 1 \\ % 
% ethnicity & Black or African American;White / Caucasian;Other. Please specify & 1 \\ % 
% ethnicity & American Indian or Alaskan Native;Hispanic & 1 \\ % 
% ethnicity & Black or African American;White / Caucasian & 1 \\ % 
% ethnicity & Hispanic;Asian / Pacific Islander & 1 \\ % 
\midrule
marital & Single & 83 \\ % 
marital & Married & 70 \\ % 
marital & Cohabitating & 26 \\ % 
marital & Divorced & 15 \\ % 
% marital & Other (Please Specify) & 4 \\ % 
% marital & Bereaved & 2 \\ % 
\midrule
language & English & 192 \\ % 
language & multi\_language & 8 \\ % 
% language & English;Other (Please Specify) & 3 \\ % 
% language & English;French & 2 \\ % 
% language & English;German & 1 \\ % 
% language & English;Arabic & 1 \\ % 
% language & English;Spanish & 1 \\ % 
\midrule
religion & Christian & 102 \\ % 
religion & Nothing & 76 \\ % 
religion & Muslim & 14 \\ % 
% religion & Other, please specify. & 4 \\ % 
% religion & Prefer not to say & 2 \\ % 
% religion & Jewish & 2 \\ % 
\midrule
home & Rent & 101 \\ % 
home & Own & 82 \\ % 
home & Other (Please specify) & 15 \\ % 
% home & Prefer not to say & 2 \\ % 
\midrule
employment & Employed full time & 102 \\ % 
employment & Employed part time & 35 \\ % 
employment & Self-employed full time & 18 \\ % 
employment & Not employed, but looking for work & 14 \\ % 
employment & Self-employed part time & 12 \\ % 
employment & Student & 6 \\ % 
% employment & Not employed, unable due to a disability or illness & 4 \\ % 
% employment & Retired & 4 \\ % 
% employment & Not employed and not looking for work & 3 \\ % 
% employment & Prefer not to say & 1 \\ % 
% employment & Stay-at-home spouse or partner & 1 \\ % 
\bottomrule
\end{tabular}
\caption{Single-value categorical attributes from the \textit{AI Gap} dataset, which we use for conditioning, sorted by their frequency (Count). We do not show attributes that appear less than 5 times. We refer to the original paper \cite{bassignana2025ai} for more details.}
\label{tab:appendix_single_value_categorical_values}
\end{center}
\end{table}

% MultiLabelBinarizer\footnote{\url{scikit-learn.org/stable/modules/generated/sklearn.preprocessing.MultiLabelBinarizer.html}}
The multi-value categorical are \textit{occupation, mother\_occupation, father\_occupation, hobbies, tech, know\_nlp, use\_nlp, would\_nlp, llm\_use, usecases, contexts}. 
Their possible values are listed in Table \ref{tab:appendix_multi_value_categorical_values}. 
For the statistical tests, they are encoded with a MultiLabelBinarizer\footnote{\url{scikit-learn.org/stable/modules/generated/sklearn.preprocessing.MultiLabelBinarizer.html}} from sklearn, and we discard attributes that appear in less than 10 different user profiles.
\begin{table*}[ht]
\scriptsize
\setlength{\tabcolsep}{2pt}
\begin{center}
\begin{tabular}{llr}
\toprule
\textbf{Attribute} & \textbf{Value} & \textbf{Count}\\
\midrule
occupation & Professionals or highly skilled workers  & 59 \\ 
occupation & Manager or business owner & 40 \\ 
occupation & Service and sales workers & 35 \\ 
occupation & Technicians and associate professionals & 28 \\
occupation & Clerical support workers & 23 \\ 
occupation & Homemaker & 9 \\ % 
occupation & Prefer not to say & 8 \\ 
occupation & Craft related trades workers  & 6 \\ 
occupation & Skilled agricultural, forestry and fishery workers & 5 \\ 
% occupation & Armed forces occupations & 2 \\ 
% occupation & Elementary occupations & 4 \\ 
% occupation & Plant and machine operators, and assemblers  & 3 \\ 
\midrule
mother\_occ. & Homemaker & 39 \\ % 
mother\_occ. & Clerical support workers  & 35 \\
mother\_occ. & Professionals or highly skilled workers & 34 \\ % 
mother\_occ. & Service and sales workers & 33 \\ % 
mother\_occ. & Manager or business owner  & 21 \\
mother\_occ. & Elementary occupations & 11 \\ % 
mother\_occ. & Technicians and associate professionals & 8 \\ % 
mother\_occ. & Skilled agricultural, forestry and fishery workers & 8 \\ % 
mother\_occ. & Craft related trades workers & 7 \\ % 
mother\_occ. & Prefer not to say & 6 \\ 
mother\_occ. & Plant and machine operators, and assemblers  & 5 \\ % 
\midrule
father\_occ. & Professionals or highly skilled workers & 46 \\ % 
father\_occ. & Manager or business owner & 40 \\ % 
father\_occ. & Service and sales workers & 27 \\ % 
father\_occ. & Craft related trades workers & 20 \\ % 
father\_occ. & Technicians and associate professionals & 16 \\ % 
father\_occ. & Prefer not to say & 13 \\ % 
father\_occ. & Plant and machine operators, and assemblers  & 11 \\ % 
father\_occ. & Elementary occupations & 11 \\ % 
father\_occ. & Clerical support workers & 9 \\ % 
father\_occ. & Armed forces occupations & 7 \\ % 
father\_occ. & Skilled agricultural, forestry and fishery workers  & 7 \\ % 
% father\_occ. & Homemaker & 2 \\ % 
\midrule
hobbies & Use social media & 141 \\ % 
hobbies & Exercise & 111 \\ % 
hobbies & Listen to rock/indie music & 73 \\ % 
hobbies & Watch sports & 73 \\ % 
hobbies & Listen to hiphop and rap & 64 \\ % 
hobbies & Do arts and crafts & 62 \\ % 
hobbies & Go to museums and galleries & 59 \\ % 
hobbies & Go to gigs & 38 \\ % 
hobbies & Listen to classical music & 37 \\ % 
hobbies & Other (Please specify) & 34 \\ % 
hobbies & Attend football matches & 30 \\ % 
hobbies & Listen to jazz & 18 \\ % 
hobbies & Watch dance or ballet & 13 \\ % 
hobbies & Visit to stately homes & 12 \\ % 
hobbies & Go to the opera & 10 \\ % 
\midrule
tech & Smartphone & 193 \\ % 
tech & Laptop & 183 \\ % 
tech & Tablet & 98 \\ % 
tech & Smartwatch & 64 \\ % 
tech & Other & 20 \\ % 
\midrule
contexts & Personal & 155 \\ % 
contexts & Work & 135 \\ % 
contexts & Learning & 131 \\ % 
contexts & Entertainment & 89 \\ % 
contexts & Creative or artistic & 72 \\ % 
contexts & School/University & 71 \\ % 
contexts & Technical & 57 \\ % 
% contexts & Other (specify) & 3 \\ % \bottomrule
\bottomrule
\end{tabular}
% \caption{First part (the rest in Table \ref{tab:appendix_multi_value_categorical_values_2}) of the multi-value categorical attributes from the \textit{AI Gap} dataset, which we use for conditioning, sorted by their frequency (Count). We do not show attributes that appear less than 5 times. The full value descriptions are shortened, we refer to the original paper \cite{bassignana2025ai} for more details.}
% \label{tab:appendix_multi_value_categorical_values_1}
% \end{center}
% \end{table}
\quad
% \begin{table}
% \scriptsize
% \setlength{\tabcolsep}{4pt}
% \begin{center}
\begin{tabular}{llr}
\toprule
\textbf{Attribute} & \textbf{Value} & \textbf{Count}\\
\midrule
llm\_use & ChatGPT & 185 \\ % 
llm\_use & Google Gemini & 99 \\ % 
llm\_use & Microsoft Bing AI & 46 \\ % 
llm\_use & GitHub Copilot & 37 \\ % 
llm\_use & Claude & 29 \\ % 
llm\_use & Snapchat My AI & 29 \\ % 
llm\_use & Character.AI & 25 \\ % 
llm\_use & Perplexity & 19 \\ % 
llm\_use & Google Bard & 19 \\ % 
llm\_use & Grok & 17 \\ % 
llm\_use & Meta Llama 2 & 14 \\ % 
llm\_use & Pi & 12 \\ % 
llm\_use & Other & 10 \\ % 
% llm\_use & HuggingChat & 2 \\ % 
llm\_use & Jasper & 6 \\ % 
llm\_use & Poe & 6 \\ % 
\midrule
know\_nlp & Spell checker & 181 \\ % 
know\_nlp & Grammar checker & 174 \\ % 
know\_nlp & E-mail spam detection & 163 \\ % 
know\_nlp & Text-to-Speech & 157 \\ % 
know\_nlp & Speech-to-Text & 156 \\ % 
know\_nlp & Question Answering \& Search Engine. & 136 \\ % 
know\_nlp & Dialog technology & 123 \\ % 
know\_nlp & Paraphrasing and Summarisation & 108 \\ % 
know\_nlp & Reading text from scanned documents & 95 \\ % 
know\_nlp & Machine translation & 85 \\ % 
know\_nlp & Sentiment analysis & 57 \\ % 
% know\_nlp & Other (specify) & 3 \\ % 
\midrule
use\_nlp & Paraphrasing and Summarisation & 167 \\ % 
use\_nlp & Grammar checker & 158 \\ % 
use\_nlp & E-mail spam detection & 142 \\ % 
use\_nlp & Question Answering \& Search Engine. & 123 \\ % 
use\_nlp & Dialog technology  & 116 \\ % 
use\_nlp & Speech-to-Text & 108 \\ % 
use\_nlp & Text-to-Speech & 99 \\ % 
use\_nlp & Machine translation  & 65 \\ % 
use\_nlp & Reading text from scanned documents & 55 \\ % 
use\_nlp & Sentiment analysis & 34 \\ % 
% use\_nlp & Other (specify) & 1 \\ % 
\midrule
would\_nlp & Sentiment analysis  & 66 \\ % 
would\_nlp & Paraphrasing and Summarisation  & 61 \\ % 
would\_nlp & Machine translation  & 59 \\ % 
would\_nlp & E-mail spam detection & 57 \\ % 
would\_nlp & Grammar checker & 54 \\ % 
would\_nlp & Speech-to-Text & 54 \\ % 
would\_nlp & Reading text from scanned documents & 54 \\ % 
would\_nlp & Text-to-Speech  & 48 \\ % 
would\_nlp & Spell checker  & 47 \\ % 
would\_nlp & Dialog technology & 41 \\ % 
would\_nlp & Question Answering \& Search Engine. & 35 \\ % 
would\_nlp & Other (specify) & 7 \\ % 
\midrule
usecases & Writing & 135 \\ % 
usecases & Brainstorming & 113 \\ % 
usecases & Answering questions about general knowledge & 110 \\ % 
usecases & Learning & 95 \\ % 
usecases & Proofreading/Editing & 95 \\ % 
usecases & Paraphrasing or finding synonyms & 88 \\ % 
usecases & Generic chatbot conversation & 85 \\ % 
usecases & Solving math or logical problems & 78 \\ % 
usecases & Translation & 75 \\ % 
usecases & Summarizing long texts & 74 \\ % 
usecases & Analyze data & 70 \\ % 
usecases & Coding & 56 \\ % 
usecases & Get information about current events & 55 \\ % 
usecases & Generate art & 46 \\ % 
usecases & Digital content creation & 46 \\ % 
usecases & Completing assignments & 39 \\ % 
usecases & Collecting references & 22 \\ % 
usecases & Play games & 14 \\ % 
% usecases & Other (specify) & 4 \\ % 
\bottomrule
\end{tabular}
\caption{List of multi-value categorical attributes from the \textit{AI Gap} dataset, which we use for conditioning, and their values (sorted by their frequency). We do not show attributes that appear less than 5 times. 
The full value names are shortened, we refer to the original paper \cite{bassignana2025ai} for more details.}
% \caption{Second part (the rest in Table \ref{tab:appendix_multi_value_categorical_values_1}). We do not show attributes that appear less than 5 times. The full value descriptions are shortened, we refer to the original paper \cite{bassignana2025ai} for more details.}
% \label{tab:appendix_multi_value_categorical_values_2}
\label{tab:appendix_multi_value_categorical_values}
\end{center}
\end{table*}

\section{Prompts}\label{sec:app_prompts}
As described in the main body of text, our prompting strategy is main of two different steps: first, we provide contextual information in explicit or implicit settings (or neither, for the \textit{model default}); then, we prompt the model to perform the task (AES, FF, QA).
The prompts used to provide contextual information are discussed in \S{}\ref{sec:app_prompts_conditioning}, and the prompts used to perform the educational task, which are agnostic to the specific conditioning, are discussed in \S{}\ref{sec:app_prompts_task}.

\subsection{Providing contextual information}\label{sec:app_prompts_conditioning}
We consider three conditions in this study: model default (\model{}), implicit demographic cue (\implicit{}), and explicit personas (\explicit{}).
For \model{}, we only give the LLM the task to perform, without any additional context.
For \implicit{}, we provide a list of previous questions asked by users from different demographic groups (according to the \textit{AI Gap} paper), and (simulated) LLM responses to those questions.
For \explicit{}, we prompt the model with an explicit description of the demographic information about the user.
This structure is shown in detail in Table \ref{tab:prompts_appendix_conditioning}. 
Please note that we only consider the last response of the model (to the last \textit{user} message) for our analysis.
\begin{table}[h]
\scriptsize
\setlength{\tabcolsep}{4pt}
\begin{center}
\begin{tabular}{lp{6.8cm}}
\toprule
% \textbf{Condition} & \textbf{Input to the LLMs} \\
\textbf{} & \textbf{Input to the LLMs} \\
\midrule
% Model default & \texttt{[\{`role': `system', `content': `'\},} \\
\multirow{6}{*}{\rotatebox[origin=c]{90}{\textbf{Model default}}} & \\
 & \\
 & \texttt{[\{`role': `system', `content': `'\},} \\
& \texttt{\{`role': `user', `content': <task\_under\_study>\},]} \\
 & \\
 & \\
\midrule
% Implicit cue &  \texttt{[{`role': `system', `content': ‘’},}\\
\multirow{9}{*}{\rotatebox[origin=c]{90}{\textbf{Implicit cue (\implicit{})}}} &  \texttt{[{`role': `system', `content': ‘’},}\\
& \texttt{{`role': `user', `content': <previous\_msg\_1>},}\\
& \texttt{{`role': `assistant', `content': <simulated\_resp\_1>},}\\
& \texttt{{`role': `user', `content': <previous\_msg\_2>},}\\
& \texttt{{`role': `assistant', `content': <simulated\_resp\_2>},}\\
& \texttt{…,}\\
& \texttt{{`role': `assistant', `content': <previous\_msg\_10>},}\\
& \texttt{{`role': `assistant', `content': <simulated\_resp\_10>},}\\
& \texttt{{`role': `user', `content': <task\_under\_study>},]}\\
\midrule
% Explicit persona & \texttt{[\{`role': `system', `content': `'\},} \\
\multirow{19}{*}{\rotatebox[origin=c]{90}{\textbf{Explicit persona (\explicit{})}}} & \texttt{[\{`role': `system', `content': `'\},} \\
& \texttt{\{`role': `user', `content': ``````}\\
& \texttt{Persona description:} \\
& \texttt{Demographics: <data> years old; nationality: <data>; ethnicity: <data>; marital status: <data>; primary language: <data>; religion: <data>.} \\
& \texttt{Education \& SES: education: <data>; SES: <data>; mother's education: <data>; father's education: <data>.} \\
& \texttt{Household \& Occupation: home setting: <data>; employment: <data>; occupation: <data>; mother's occupation: <data>; father's occupation: <data>.} \\
& \texttt{Technology \& NLP: general tech familiarity: <data>; NLP knowledge: <data>; past NLP use: <data>; willingness to use NLP tools: <data>; LLM usage frequency: <data>; LLM typical use: <data>.} \\
& \texttt{Interests \& Hobbies: <data>.} \\
& \texttt{NLP/AI Use-Cases of Interest: <data>.} \\
& \texttt{Usage Contexts: <data>.} \\
& \texttt{'''''' + <task\_under\_study>\},]} \\
\bottomrule
\end{tabular}
\caption{Structure of the messages sent to the LLMs for the three conditions under study. In \implicit{}, the \texttt{<previous\_msg\_*>} are the prompt histories from the \textit{AI Gap} paper, and the \texttt{<simulated\_resp\_*>} are the responses provided by GPT-5-nano. In \explicit{}, the \texttt{<data>} are the explicit demographic information taken from the \textit{AI Gap} paper. The \texttt{<task\_under\_study>} is the part of the prompt that describes the educational task to perform (see Table \ref{tab:prompt_structure_tasks}).}
\label{tab:prompts_appendix_conditioning}
\end{center}
\end{table}

\subsection{Performing the educational tasks}\label{sec:app_prompts_task}
While the prompts differ depending on the task under consideration (AES, FF, or QA), the structure is shared across them.
The structure of \texttt{<task\_under\_study>} differs between the three tasks under consideration (AES, FF, QA).
For Automated Essay Scoring (AES), we ask the model to rate essay based on the overall writing quality, where 6 indicates clear and consistent mastery. 
For the Formative Feedback task (FF), we prompt the model to provide a brief summary of the strengths and weaknesses of each essay.
For the metalinguistic QA task, the prompt is the user's question on the Stack Exchange forum, with no additional information.
Table \ref{tab:prompt_structure_tasks} shows the structure for the three tasks.
\begin{table}[t]
\scriptsize
\centering
\begin{tabularx}{\columnwidth}{lX}
\toprule
\textbf{Task} & \textbf{Description} \\
\midrule
AES \& FF &
\raggedright\arraybackslash
\texttt{After reading each essay and completing the analytical rating form, assign a holistic score based on the rubric below. Use a grading scale from 1 (minimum) to 6 (maximum), where the distance between adjacent scores is considered equal.}\\[0.3em]
&
\raggedright\arraybackslash
\texttt{Score 1: very little or no mastery; severely flawed in argumentation, organization, language use, and grammar.}\\
&
\raggedright\arraybackslash
\texttt{Score 2: little mastery; vague position, weak evidence, poor organization, limited language facility.}\\
&
\raggedright\arraybackslash
\texttt{Score 3: developing mastery; some critical thinking but inconsistent support and organization.}\\
&
\raggedright\arraybackslash
\texttt{Score 4: adequate mastery; competent reasoning, adequate support, generally organized writing.}\\
&
\raggedright\arraybackslash
\texttt{Score 5: reasonably consistent mastery; strong critical thinking, coherent organization, effective language use.}\\
&
\raggedright\arraybackslash
\texttt{Score 6: clear and consistent mastery; insightful reasoning, strong evidence, skillful language use.}\\[0.3em]
&
\raggedright\arraybackslash
\texttt{Writing Assignment: \{writing\_assignment\}}\\[0.3em]
&
\raggedright\arraybackslash
\texttt{Source Text: \{source\_text\}}\\[0.3em]
&
\raggedright\arraybackslash
\texttt{Essay: \{essay\_text\}}\\[0.3em]
&
\raggedright\arraybackslash
\texttt{\{"holistic\_essay\_score": <integer 1--6>, "summary": "brief summary of strengths and weaknesses"\}} \\
\midrule
QA &
\texttt{\{metalinguistic\_question\}} \\
\bottomrule

\end{tabularx}
\caption{Texts used to instruct the LLMs to perform the educational tasks. For the AES task, we use the scoring instructions provided in the \href{PERSUADE 2.0 dataset}{PERSUADE 2.0 dataset}. The placeholders \texttt{\{essay\_text\}}, \texttt{\{source\_text\}}, \texttt{\{writting\_assignment\}} and \texttt{\{metalinguistic\_question\}} are replaced with examples from the datasets.}
\label{tab:prompt_structure_tasks}
\end{table}

% Texts used to instruct the LLMs to perform the educational tasks. For the AES task, we use the scoring instructions provided in the PERSUADE 2.0 dataset\footnote{\url{https://github.com/scrosseye/persuade_corpus_2.0/blob/main/sat_rubric_only_source_based.pdf}}. The placeholders \texttt{\{source\_text\}}, \texttt{\{essay\_text\}}, \texttt{\{writing\_assignment\}}, and \texttt{\{metalinguistic\_question\}} are replaced with instances from the corresponding datasets.

\section{Libraries and Packages}
\label{sec:app:pkg}
\subsection{Linguistic Metrics}
\label{sec:app:pkg-metrics}
\begin{itemize}
    \item \textit{ARI} and \textit{FRE} computed via \texttt{textstat} package.\footnote{ \url{https://pypi.org/project/textstat/}} .
    
    \item \textit{Sentiment}: We calculate the sentiment of the response, measured with the BERT-based, education-fine-tuned sentiment classifier\footnote{Available on \url{https://huggingface.co/AventIQ-AI/sentiment-analysis-for-student-feedback-analysis}}.
    
    \item \textit{BERTScore}: We use the \texttt{score} method from the \texttt{bert\_score} package and consider the BERTScore F1 score in the analysis.\footnote{\url{https://pypi.org/project/bert-score/}}
\end{itemize}

\subsection{Statistical Tests}
\label{sec:app:pkg-st}
\begin{itemize}
    \item \textit{OLS model}: We use \texttt{statsmodels} package.\footnote{ \url{hhttps://www.statsmodels.org/stable/api.html}} 
    \item \textit{KS test}: We use \texttt{ks\_2samp} from \texttt{scipy.stats} package.
    \item \textit{Bonferroni correction}: Implemented with \texttt{statsmodels}\footnote{\url{statsmodels.org/dev/generated/statsmodels.stats.multitest.multipletests.html}}

\end{itemize}

\section{Models}\label{app:exp_setup}
Table \ref{tab:models_used} lists the models that we use for the experiments described in this paper. We query the GPT models via the OpenAI API and prompt the open-weight models from HuggingFace after deploying them on local resources (using \texttt{vLLM}). For all models, we use \texttt{temperature = 0.0}; for Llama 3.3 70B and Qwen3 30B we use \texttt{bitsandbytes} quantization as provided by vLLM, and limit the context length to 32,768 tokens (we do \textit{not} use this hard limit for the smaller Llama 3.2 7B and Qwen3 4B, nor for the GPT models).

\begin{table}[ht]
\centering
\scriptsize
\setlength{\tabcolsep}{4pt}
\begin{tabular}{ll}
\toprule
\textbf{Model} & \textbf{Identifier} \\
\midrule
GPT-5-nano & \texttt{gpt-5-nano-2025-08-07} \\
GPT-5-mini & \texttt{gpt-5-mini-2025-08-07} \\
\midrule
Llama-3B & \texttt{meta-llama/Llama-3.2-3B-Instruct} \\
Llama-70B & \texttt{meta-llama/Llama-3.3-70B-Instruct} \\
Qwen-4B & \texttt{Qwen/Qwen-4B-Instruct-2507} \\
Qwen-30B & \texttt{Qwen/Qwen-30B-A3B-Instruct-2507} \\
\bottomrule
\end{tabular}
\caption{List of models used for the experiments.}
\label{tab:models_used}
\end{table}

\section{Additional Results}
\subsection{Topic Modeling on User Prompts}
\label{sec:app:tm}

\begin{figure*}
    \centering
    \includegraphics[width=1\linewidth]{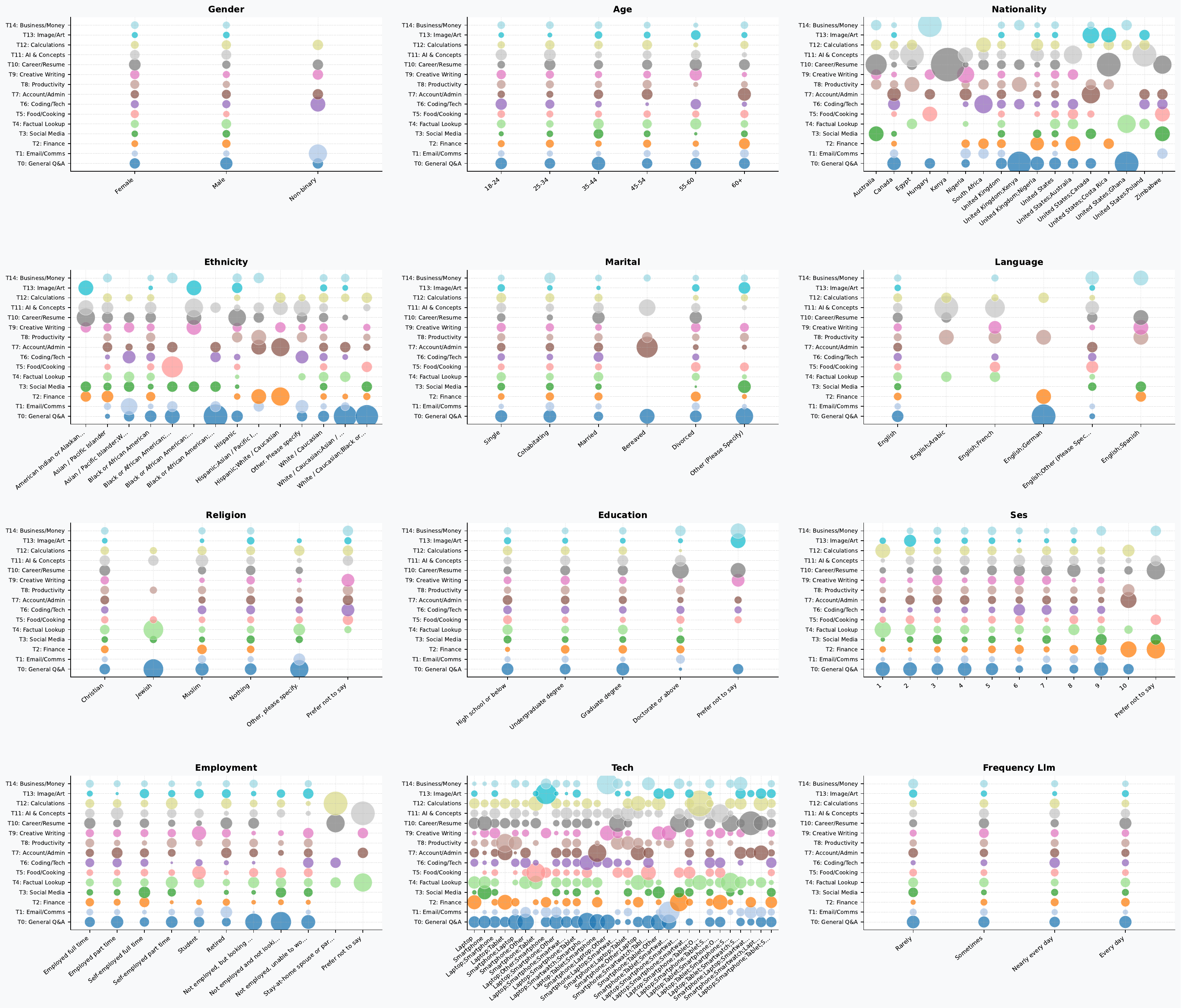}
    \caption{Topic distribution across demographic groups derived from KMeans clustering (k = 15) applied to 2,000 simulated user prompts (200 participants, 10 prompts each). Each bubble represents the share of prompts assigned to a given topic (y-axis) within a demographic subgroup (x-axis); bubble size is proportional to that percentage. Groups with fewer than 5 users are excluded.}
    \label{fig:topic-model}
\end{figure*}

Topic labels were derived post-hoc using TF-IDF top-10 terms per cluster. All topics and related terms are presented in \Cref{tab:topic-list}. Across 15 topics, the most populated were \textit{General Q\&A} (T0, n = 240), \textit{Career \& Resume} (T10, n = 185), and \textit{AI \& Concept Explanation} (T11, n = 179). The results of the topic model are presented in \Cref{fig:topic-model}. Demographic breakdowns revealed several patterns: \textbf{Gender}: Women's prompts skew toward career/resume tasks (T10, 12.5\%); men's toward general Q\&A (T0, 13.8\%). \textbf{Age}: Younger users (25–34) show the highest share of AI/concept prompts (T11); older users (45–54, 60+) cluster on account/admin tasks (T7). \textbf{SES}: lower SES (1) dominates factual lookup (T4); mid-high SES (6–7) shifts to coding and AI explanations; top SES (10) shows the strongest finance signal (T2, 23\%). \textbf{Employment}: Full-time employees favour career/resume (T10); students favour creative writing (T9); self-employed users lean toward social media content (T3). \textbf{LLM frequency}: Daily users default to general Q\&A; "nearly every day" users are the most AI/concept-oriented subgroup.

\begin{table*}[ht]
\centering
\scriptsize
\setlength{\tabcolsep}{4pt}
\begin{tabular}{clrp{9.5cm}}
\toprule
\textbf{Topic} & \textbf{Label} & \textbf{n} & \textbf{Top TF-IDF terms} \\
\midrule
T0  & General Q\&A / Health        & 240 & does, think, like, signs, causes, tall, people, know, dexter, time \\
T1  & Email writing                & 80  & email, write, translate, make, help, professional, want, sound, draft, good \\
T2  & Finance \& stocks            & 112 & stock, stocks, market, financial, uk, explain, tax, invest, money, company \\
T3  & Media \& social content      & 98  & video, youtube, tv, description, title, media, good, rain, channel, games \\
T4  & Factual lookup               & 152 & weather, did, win, today, time, year, 2025, does, usa, tomorrow \\
T5  & Food \& recipes               & 101 & make, recipe, dinner, food, best, soup, restaurants, ingredients, does, meal \\
T6  & Coding                       & 126 & function, create, js, make, way, best, button, php, excel, learn \\
T7  & Account \& admin             & 144 & account, use, need, access, uk, does, data, information, phone, help \\
T8  & Health \& productivity       & 124 & best, work, time, ways, strategies, improve, stress, productivity, day, workout \\
T9  & Creative writing             & 129 & write, story, short, poem, help, tell, summary, make, books, like \\
T10 & Career \& resume             & 185 & write, customer, resume, help, service, skills, job, best, business, questions \\
T11 & AI \& concept explanation    & 179 & ai, explain, difference, terms, simple, learning, research, concept, data, tell \\
T12 & Calculations                 & 146 & calculate, number, cost, month, 100, day, 12, miles, population, tell \\
T13 & Image generation             & 79  & image, make, create, art, picture, creation, cat, cute, ninja, like \\
T14 & Business ideas               & 105 & best, money, good, make, new, business, start, want, selling, things \\
\bottomrule
\end{tabular}
\caption{Topics identified by KMeans clustering ($k=15$) on 2{,}000 simulated user prompts, with post-hoc labels and top-10 TF-IDF terms per cluster. Only T0, T10, and T11 are named in the main text; remaining labels are descriptive glosses derived from top terms and example prompts.}
\label{tab:topic-list}
\end{table*}

\subsection{Automated Essay Scoring}
\label{sec:app:aes}
\begin{table}[ht]
\centering
\scriptsize
\begin{tabularx}{\columnwidth}{l X c r}
\toprule
\textbf{Model} & \textbf{Attribute} & $\mu{}_b\;(\sigma{}_b)$ & \textbf{Cnt} \\
\midrule
  \textbf{Qwen-30B} & \textbf{education} & $\mathbf{-0.11}$ $\mathbf{(0.08)}$ & \textbf{12} \\
  Qwen-4B & frequency\_llm & $+0.04$ $(0.07)$ & 15 \\
  Qwen-30B & mum\_education & $-0.03$ $(0.02)$ & 11 \\
  Qwen-4B & education & $-0.02$ $(0.09)$ & 14 \\
  Qwen-4B & age & $-0.01$ $(0.06)$ & 11 \\
  Qwen-4B & dad\_education & $-0.01$ $(0.03)$ & 11 \\
  GPT-mini & ses & $+0.00$ $(0.04)$ & 12 \\
  Llama-3B & education & $+0.00$ $(0.11)$ & 13 \\
  Llama-3B & frequency\_llm & $+0.00$ $(0.03)$ & 12 \\
\midrule
  \textbf{Llama-70B} & \textbf{know\_nlp: Text-to-Speech } & $\mathbf{+0.27}$ $\mathbf{(0.10)}$ & \textbf{5} \\
  \textbf{Llama-70B} & \textbf{usecases: Brainstorming} & $\mathbf{+0.26}$ $\mathbf{(0.16)}$ & \textbf{24} \\
  \textbf{Llama-70B} & \textbf{usecases: Writing} & $\mathbf{+0.24}$ $\mathbf{(0.08)}$ & \textbf{19} \\
  \textbf{Llama-70B} & \textbf{tech: Laptop} & $\mathbf{+0.21}$ $\mathbf{(0.05)}$ & \textbf{5} \\
  \textbf{Llama-70B} & \textbf{religion: Nothing} & $\mathbf{+0.19}$ $\mathbf{(0.12)}$ & \textbf{20} \\
  \textbf{Llama-70B} & \textbf{hobbies: arts} & $\mathbf{+0.18}$ $\mathbf{(0.13)}$ & \textbf{15} \\
  \textbf{Llama-70B} & \textbf{would\_nlp: Dialog tech.} & $\mathbf{+0.17}$ $\mathbf{(0.12)}$ & \textbf{20} \\
  \textbf{Llama-70B} & \textbf{hobbies: museums} & $\mathbf{+0.16}$ $\mathbf{(0.11)}$ & \textbf{6} \\
  \textbf{Llama-70B} & \textbf{mum\_occupation: Service and sales workers} & $\mathbf{+0.16}$ $\mathbf{(0.07)}$ & \textbf{8} \\
  \textbf{Llama-70B} & \textbf{llm\_use: GitHub Copilot} & $\mathbf{+0.12}$ $\mathbf{(0.08)}$ & \textbf{7} \\
  \textbf{Llama-70B} & \textbf{gender: Female} & $\mathbf{+0.12}$ $\mathbf{(0.08)}$ & \textbf{21} \\
  \textbf{Llama-70B} & \textbf{language: English} & $\mathbf{+0.10}$ $\mathbf{(0.04)}$ & \textbf{13} \\
  \textbf{Llama-70B} & \textbf{use\_nlp: Dialog tech.} & $\mathbf{+0.09}$ $\mathbf{(0.05)}$ & \textbf{8} \\
  \textbf{Llama-70B} & \textbf{religion: Christian} & $\mathbf{+0.09}$ $\mathbf{(0.07)}$ & \textbf{6} \\
  Llama-70B & would\_nlp: Search Eng. & $+0.08$ $(0.05)$ & 13 \\
  Llama-3B & language: English & $+0.08$ $(0.07)$ & 10 \\
  Qwen-30B & usecases: Writing & $+0.08$ $(0.04)$ & 6 \\
  Llama-70B & employment: Employed full time & $+0.07$ $(0.05)$ & 17 \\
  Llama-70B & ses & $+0.07$ $(0.05)$ & 17 \\
  Llama-70B & usecases: Solving math or logical problems & $-0.05$ $(0.05)$ & 9 \\
  Qwen-30B & religion: Nothing & $+0.05$ $(0.06)$ & 7 \\
  Llama-3B & dad\_education & $-0.04$ $(0.03)$ & 7 \\
  Llama-70B & mum\_education & $+0.03$ $(0.02)$ & 13 \\
  Llama-70B & education & $+0.03$ $(0.03)$ & 19 \\
  Llama-70B & dad\_education & $-0.03$ $(0.02)$ & 7 \\
  Qwen-30B & religion: Christian & $+0.02$ $(0.03)$ & 6 \\
  GPT-nano & education & $-0.02$ $(0.03)$ & 7 \\
  Qwen-30B & ses & $+0.02$ $(0.03)$ & 9 \\
  Llama-70B & age & $-0.01$ $(0.04)$ & 11 \\
  Llama-3B & frequency\_llm & $-0.01$ $(0.04)$ & 7 \\
  Qwen-30B & usecases: Solving math or logical problems & $+0.01$ $(0.03)$ & 5 \\
  Qwen-30B & education & $-0.01$ $(0.02)$ & 8 \\
  Qwen-30B & mum\_education & $-0.01$ $(0.02)$ & 5 \\
  Llama-3B & ses & $+0.01$ $(0.03)$ & 6 \\
  Qwen-4B & education & $+0.00$ $(0.02)$ & 5 \\
  Qwen-30B & nationality: United States & $+0.00$ $(0.04)$ & 7 \\
  Qwen-4B & nationality: United States & $+0.00$ $(0.03)$ & 5 \\
  Qwen-30B & dad\_education & $+0.00$ $(0.02)$ & 6 \\
  Llama-3B & age & $+0.00$ $(0.01)$ & 5 \\
\bottomrule
\end{tabularx}
\caption{Impact of different demographic attributes in \explicit{} (top) and \implicit{} (bottom) on the score of the model responses, ordered by $|\mu_b|$. In bold if $|\mu_b| \geq \sigma_\text{resp.score}$}
\label{tab:results_regression_test_aes_all}
\end{table}

In \Cref{tab:results_regression_test_aes_all} we show all results related to the regression test over the AES task. The KS test results in \Cref{tab:ks-aes} show that demographic-related shifts in AES score deviation occur only under the \explicit{} persona condition and not under \implicit{} conditioning. 

\begin{table}[ht]
\centering
\scriptsize
\begin{tabular}{llllcr}
\toprule
\textbf{Model} & \textbf{Attribute} & \textbf{V.1} & \textbf{V.2} & $\mu{}_D$ ($\sigma{}_D$) & \textbf{Cnt} \\
\midrule
  Llama-3B & education & 2 & 0 & ${0.65}$ ${(0.12)}$ & 4 \\
  Qwen-30B & education & 2 & 0 & ${0.60}$ ${(0.14)}$ & 7 \\
  Qwen-4B & frequency\_llm & 3 & 0 & ${0.59}$ ${(0.09)}$ & 4 \\
  Qwen-4B & frequency\_llm & 2 & 0 & ${0.57}$ ${(0.11)}$ & 3 \\
  Llama-3B & education & 0 & 1 & $0.48$ $(0.09)$ & 5 \\
  Qwen-30B & education & 0 & 1 & $0.44$ $(0.03)$ & 3 \\
  Qwen-4B & frequency\_llm & 3 & 1 & $0.41$ $(0.03)$ & 3 \\
  Llama-3B & education & 2 & 1 & $0.40$ $(0.02)$ & 2 \\
  Qwen-30B & education & 2 & 1 & $0.39$ $(0.04)$ & 2 \\
\midrule
  \multicolumn{6}{l}{\textit{No significant results (Implicit).}} \\
\bottomrule
\end{tabular}
\caption{Impact of different demographic attributes in \explicit{} (top) and \implicit{} (bottom) conditions on the AES score deviation, according to the KS test, ordered by average magnitude of $\mu_D$ (the KS statistic $D$). V.1/V.2 = ordinal codes for ordered attributes, category labels otherwise. $\mu_D$ = mean $D$ across essays where the pair was significant; $\sigma_D$ = std; Cnt = number of essays of 40.}
\label{tab:ks-aes}
\end{table}

  \begin{table*}[h]
\centering
\scriptsize
\begin{tabular}{llllcr}
\toprule
\textbf{Model} & \textbf{Attribute} & \textbf{V.1} & \textbf{V.2} & $\mu{}_D$ ($\sigma{}_D$) & \textbf{Cnt} \\
\midrule
  Llama-3B & father\_occupation & Prefer not to say & Technician/associate & ${0.79}$ & 1 \\
  Llama-70B & ses & 2 & 4 & ${0.78}$ & 1 \\
  Llama-70B & ses & 2 & 8 & ${0.76}$ & 1 \\
  Llama-3B & father\_occupation & Elementary occupation & Manager/business owner & ${0.71}$ & 1 \\
  Llama-3B & father\_occupation & Prefer not to say & Professional/skilled worker & ${0.71}$ ${(0.08)}$ & 3 \\
  Llama-70B & ses & 2 & 3 & ${0.67}$ & 1 \\
  Llama-70B & ses & 2 & 7 & ${0.64}$ ${(0.03)}$ & 3 \\
  Llama-70B & ses & 3 & 8 & ${0.62}$ & 1 \\
  Llama-70B & ses & 2 & 5 & ${0.61}$ & 1 \\
  Llama-70B & ses & 2 & 6 & ${0.60}$ & 1 \\
  Llama-3B & occupation & Manager/business owner & Technician/associate & ${0.59}$ & 1 \\
  Llama-70B & ses & 3 & 4 & ${0.57}$ & 1 \\
  Llama-3B & occupation & Professional/skilled worker & Technician/associate & ${0.57}$ ${(0.02)}$ & 2 \\
  Llama-3B & father\_occupation & Craft/trades worker & Professional/skilled worker & ${0.56}$ & 1 \\
  Llama-3B & occupation & Manager/business owner & Service/sales worker & ${0.54}$ & 1 \\
  Llama-3B & father\_occupation & Professional/skilled worker & Service/sales worker & ${0.53}$ & 1 \\
  Llama-70B & ses & 3 & 7 & ${0.53}$ ${(0.00)}$ & 2 \\
  Llama-3B & father\_occupation & Manager/business owner & Professional/skilled worker & ${0.52}$ & 1 \\
  Llama-3B & occupation & Professional/skilled worker & Service/sales worker & ${0.52}$ ${(0.04)}$ & 5 \\
  Qwen-4B & education & 2 & 0 & $0.48$ $(0.08)$ & 4 \\
  Llama-70B & education & 2 & 0 & $0.48$ $(0.08)$ & 10 \\
  Llama-70B & education & 0 & 1 & $0.47$ $(0.11)$ & 6 \\
  Llama-3B & education & 2 & 0 & $0.47$ $(0.09)$ & 11 \\
  Llama-3B & education & 0 & 1 & $0.42$ $(0.06)$ & 10 \\
  Llama-3B & education & 2 & 1 & $0.42$ & 1 \\
  Llama-3B & nationality & UK & US & $0.41$ $(0.10)$ & 12 \\
  Qwen-30B & education & 2 & 0 & $0.40$ $(0.02)$ & 4 \\
  Llama-70B & nationality & UK & US & $0.39$ $(0.08)$ & 8 \\
  Qwen-30B & education & 0 & 1 & $0.39$ $(0.05)$ & 3 \\
  Qwen-4B & education & 0 & 1 & $0.39$ $(0.02)$ & 3 \\
  Qwen-4B & nationality & UK & US & $0.38$ $(0.06)$ & 5 \\
\midrule
  \multicolumn{6}{l}{\textit{No significant results (Implicit).}} \\
\bottomrule
\end{tabular}
\caption{Impact of different demographic attributes in \explicit{} (top) and \implicit{} (bottom) conditions on \textit{Age of Acquisition ($\Delta$)}, according to the KS test, ordered by average magnitude of $\mu_D$ (the KS statistic $D$). V.1/V.2 = ordinal codes for ordered attributes, category labels otherwise. $\mu_D$ = mean $D$ across essays where the pair was significant; $\sigma_D$ = std; Cnt = number of essays of 40.}
\label{tab:ks-ff-aoa}
\end{table*}

\subsection{Formative Feedback}\label{sec:app:ff}
\paragraph{Age of Acquisition (AoA)}\label{sec:app:ff:aoa}
\begin{table}[ht]
\centering
\scriptsize
\begin{tabularx}{\columnwidth}{l X c r}
\toprule
\textbf{Model} & \textbf{Attribute} & $\mu{}_b\;(\sigma{}_b)$ & \textbf{Cnt} \\
\midrule
  \textbf{Qwen-4B} & \textbf{tech: Laptop} & $\mathbf{+0.03}$ $\mathbf{(0.03)}$ & \textbf{10} \\
\midrule
  \textbf{Llama-70B} & \textbf{language: English} & $\mathbf{+0.03}$ $\mathbf{(0.02)}$ & \textbf{8} \\
  \textbf{Qwen-30B} & \textbf{language: English} & $\mathbf{+0.02}$ $\mathbf{(0.02)}$ & \textbf{7} \\
  \textbf{GPT-mini} & \textbf{language: English} & $\mathbf{+0.02}$ $\mathbf{(0.01)}$ & \textbf{5} \\
\bottomrule
\end{tabularx}
\caption{Impact of different demographic attributes in \explicit{} (top) and \implicit{} (bottom) on the AoA of the model responses (using \model{} as reference), ordered by $|\mu_b|$ in the FF task. In bold if $|\mu_b| \geq \sigma_\text{AoA}$.}
\label{tab:ff-ols-aoa}
\end{table}

Qwen-4B shows significant results with profiles related to the \textit{tech:Laptop} attribute in the \explicit{} condition (\Cref{tab:ff-ols-aoa}). In the KS test, the largest distributional differences involve father's occupation in Llama-3B, where ``Prefer not to say'' vs. ``technician/associate'' has D=0.79. SES shows large gaps in Llama-70B (e.g., ses2 vs. ses4: D=0.78). Education shows consistent effects across Llama-3B, Llama-70B, Qwen-30B, and Qwen-4B, with larger gaps between more distant education levels (e.g., edu2 vs. edu0 > edu1 vs. edu0). Nationality (UK vs. US) is significant for Llama-3B, Llama-70B, and Qwen-4B (\Cref{tab:ks-ff-aoa}).

\paragraph{Flesch Reading Ease (FRE)  }\label{sec:app:ff:fre}
\begin{table}[ht]
\centering
\scriptsize
\begin{tabularx}{\columnwidth}{l X c r}
\toprule
\textbf{Model} & \textbf{Attribute} & $\mu{}_b\;(\sigma{}_b)$ & \textbf{Cnt} \\
\midrule
  \multicolumn{4}{l}{\textit{No rows pass filters (Explicit).}} \\
\midrule
  \textbf{Llama-70B} & \textbf{usecases: Writing} & $\mathbf{+3.67}$ $\mathbf{(1.05)}$ & \textbf{6} \\
  \textbf{Llama-70B} & \textbf{know\_nlp: spam detec} & $\mathbf{+1.95}$ $\mathbf{(1.14)}$ & \textbf{6} \\
  \textbf{Llama-70B} & \textbf{hobbies: museums} & $\mathbf{+1.93}$ $\mathbf{(1.87)}$ & \textbf{5} \\
  \textbf{Llama-70B} & \textbf{hobbies: Exercise} & $\mathbf{+1.72}$ $\mathbf{(0.82)}$ & \textbf{5} \\
  \textbf{Llama-70B} & \textbf{religion: Christian} & $\mathbf{+1.41}$ $\mathbf{(1.15)}$ & \textbf{6} \\
  \textbf{GPT-nano} & \textbf{language: English} & $\mathbf{+1.35}$ $\mathbf{(0.88)}$ & \textbf{8} \\
  \textbf{Llama-70B} & \textbf{language: English} & $\mathbf{+1.19}$ $\mathbf{(1.12)}$ & \textbf{5} \\
  \textbf{Llama-70B} & \textbf{home: Own} & $\mathbf{+1.12}$ $\mathbf{(0.74)}$ & \textbf{5} \\
  \textbf{Llama-70B} & \textbf{gender: Female} & $\mathbf{+1.09}$ $\mathbf{(0.65)}$ & \textbf{6} \\
  \textbf{GPT-mini} & \textbf{language: English} & $\mathbf{+0.87}$ $\mathbf{(0.50)}$ & \textbf{5} \\
  \textbf{Qwen3-30B} & \textbf{language: English} & $\mathbf{+0.86}$ $\mathbf{(0.36)}$ & \textbf{5} \\
\bottomrule
\end{tabularx}
\caption{Impact of different demographic attributes in \explicit{} (top) and \implicit{} (bottom) on the FRE of the model responses (using \model{} as reference), ordered by $|\mu_b|$ in the FF task. In bold if $|\mu_b| \geq \sigma_\text{FRE}$.}
\label{tab:ff-ols-fre}
\end{table}

Under \implicit{} personas, 11 significant predictors emerge, entirely positive, which means greater readability compared to the model default (\Cref{tab:ff-ols-fre}). Llama-70B shows the largest and most varied set of predictors: persona profiles associated with writing email use cases ($\mu_b = +3.67$, Cnt=6), hobbies of visiting museums ($+1.93$), exercising ($+1.72$), Christian religion ($+1.41$), English language ($+1.19$), home ownership ($+1.12$), and female gender ($+1.09$) all receive more readable feedback. GPT-nano and GPT-mini also show significant language: English effects ($+1.35$ and $+0.87$), as does Qwen-30B ($+0.86$). We do not observe any significant results under \explicit{}.
\begin{table*}[ht]
\centering
\scriptsize
\begin{tabular}{llllcr}
\toprule
\textbf{Model} & \textbf{Attribute} & \textbf{V.1} & \textbf{V.2} & $\mu{}_D$ ($\sigma{}_D$) & \textbf{Cnt} \\
\midrule
  Llama-3B & ses & 4 & 7 & ${0.66}$ & 1 \\
  Llama-70B & occupation & Clerical worker & Manager/business owner & ${0.63}$ & 1 \\
  Llama-3B & occupation & Service/sales worker & Technician/associate & ${0.63}$ & 1 \\
  Qwen-4B & occupation & Clerical worker & Professional/skilled worker & ${0.62}$ ${(0.05)}$ & 2 \\
  Llama-3B & ses & 3 & 4 & ${0.61}$ & 1 \\
  Llama-3B & ses & 2 & 3 & ${0.61}$ & 1 \\
  Llama-3B & ses & 2 & 6 & ${0.60}$ & 1 \\
  Llama-3B & occupation & Manager/business owner & Technician/associate & ${0.59}$ & 1 \\
  Llama-3B & ses & 4 & 6 & ${0.58}$ & 1 \\
  Llama-70B & occupation & Clerical worker & Professional/skilled worker & ${0.58}$ & 1 \\
  Llama-3B & occupation & Professional/skilled worker & Service/sales worker & ${0.55}$ ${(0.06)}$ & 4 \\
  Llama-70B & education & 2 & 0 & ${0.55}$ ${(0.08)}$ & 6 \\
  Llama-3B & ses & 5 & 6 & ${0.55}$ & 1 \\
  Llama-3B & ses & 6 & 7 & ${0.55}$ & 1 \\
  Llama-3B & ses & 4 & 5 & ${0.55}$ & 1 \\
  Llama-3B & occupation & Professional/skilled worker & Technician/associate & ${0.54}$ & 1 \\
  Llama-3B & ses & 3 & 7 & ${0.53}$ ${(0.01)}$ & 2 \\
  Llama-70B & occupation & Professional/skilled worker & Service/sales worker & ${0.52}$ ${(0.08)}$ & 4 \\
  Qwen-4B & occupation & Manager/business owner & Service/sales worker & ${0.51}$ & 1 \\
  Llama-70B & education & 0 & 1 & $0.50$ $(0.13)$ & 6 \\
  Qwen-4B & occupation & Professional/skilled worker & Service/sales worker & $0.49$ $(0.04)$ & 2 \\
  Llama-3B & education & 2 & 0 & $0.46$ $(0.07)$ & 10 \\
  Llama-3B & nationality & UK & US & $0.42$ $(0.13)$ & 11 \\
  Qwen-4B & nationality & UK & US & $0.41$ $(0.13)$ & 5 \\
  Llama-3B & education & 0 & 1 & $0.40$ $(0.05)$ & 9 \\
  Llama-3B & education & 2 & 1 & $0.40$ & 1 \\
  Llama-70B & nationality & UK & US & $0.37$ $(0.06)$ & 7 \\
\midrule
  \multicolumn{6}{l}{\textit{No significant results (Implicit).}} \\
\bottomrule
\end{tabular}
\caption{Impact of different demographic attributes in \explicit{} (top) and \implicit{} (bottom) conditions on \textit{Flesch Reading Ease ($\Delta$)}, according to the KS test, ordered by average magnitude of $\mu_D$ (the KS statistic $D$). V.1/V.2 = ordinal codes for ordered attributes, category labels otherwise. $\mu_D$ = mean $D$ across essays where the pair was significant; $\sigma_D$ = std; Cnt = number of essays of 40.}
\label{tab:ks-ff-fri}
\end{table*}

However, with the KS test as presented in \Cref{tab:ks-ff-fri}, only \explicit{} conditioning shows significant results. We observe SES in Llama-3B, with the largest D values reaching 0.66 (Llama-3B, ses4 vs. ses7). Occupation and education affect the outputs of Qwen-4B, Llama-3B, and Llama-70B. For example, when the occupation is explicitly defined as ``Technician/associate'', Llama-3B significantly increases FRE compared with other occupations. We find Nationality as a significant effect for Llama-3B, Llama-70B, and Qwen-4B. It indicates that US nationality vs UK in \explicit{}, the generated feedback has higher readability.

\begin{table}[ht]
\centering
\scriptsize
\begin{tabularx}{\columnwidth}{l X c r}
\toprule
\textbf{Model} & \textbf{Attribute} & $\mu{}_b\;(\sigma{}_b)$ & \textbf{Cnt} \\
\midrule
  \textbf{GPT-nano} & \textbf{language: English} & $\mathbf{+0.46}$ $\mathbf{(0.31)}$ & \textbf{11} \\
\midrule
  \textbf{Llama-70B} & \textbf{hobbies: Use social media} & $\mathbf{+0.73}$ $\mathbf{(0.46)}$ & \textbf{5} \\
  \textbf{Llama-70B} & \textbf{hobbies: museums} & $\mathbf{+0.56}$ $\mathbf{(0.25)}$ & \textbf{8} \\
  \textbf{GPT-nano} & \textbf{hobbies: Exercise} & $\mathbf{+0.32}$ $\mathbf{(0.19)}$ & \textbf{5} \\
\bottomrule
\end{tabularx}
\caption{Impact of different demographic attributes in \explicit{} (top) and \implicit{} (bottom) on the ARI of the model responses (using \model{} as reference), ordered by $|\mu_b|$ in the FF task. In bold if $|\mu_b| \geq \sigma_\text{ARI}$.}
\label{tab:ff-ols-ari}
\end{table}

\begin{table*}[h!]
\centering
\scriptsize
\begin{tabular}{llllcr}
\toprule
\textbf{Model} & \textbf{Attribute} & \textbf{V.1} & \textbf{V.2} & $\mu{}_D$ ($\sigma{}_D$) & \textbf{Cnt} \\
\midrule
  Llama-3B & father\_occupation & Prefer not to say & Technician/associate & ${0.78}$ ${(0.00)}$ & 2 \\
  Llama-3B & father\_occupation & Craft/trades worker & Prefer not to say & ${0.76}$ & 1 \\
  Llama-3B & father\_occupation & Craft/trades worker & Technician/associate & ${0.71}$ & 1 \\
  Llama-3B & father\_occupation & Prefer not to say & Professional/skilled worker & ${0.66}$ ${(0.02)}$ & 4 \\
  Llama-3B & occupation & Manager/business owner & Technician/associate & ${0.61}$ & 1 \\
  Llama-3B & occupation & Service/sales worker & Technician/associate & ${0.60}$ ${(0.02)}$ & 2 \\
  Llama-3B & father\_occupation & Craft/trades worker & Professional/skilled worker & ${0.58}$ & 1 \\
  Llama-3B & occupation & Professional/skilled worker & Technician/associate & ${0.55}$ ${(0.03)}$ & 3 \\
  Llama-3B & occupation & Professional/skilled worker & Service/sales worker & ${0.54}$ ${(0.04)}$ & 5 \\
  Llama-70B & education & 2 & 0 & ${0.50}$ ${(0.06)}$ & 7 \\
  Llama-70B & education & 0 & 1 & ${0.50}$ ${(0.13)}$ & 4 \\
  Llama-3B & education & 2 & 0 & $0.47$ $(0.07)$ & 6 \\
  Llama-3B & nationality & UK & US & $0.44$ $(0.13)$ & 11 \\
  Qwen-4B & education & 2 & 0 & $0.43$ $(0.05)$ & 5 \\
  Llama-3B & education & 0 & 1 & $0.43$ $(0.06)$ & 7 \\
  Qwen-4B & nationality & UK & US & $0.39$ $(0.03)$ & 8 \\
  Qwen-4B & education & 0 & 1 & $0.38$ $(0.01)$ & 2 \\
  Llama-70B & nationality & UK & US & $0.36$ $(0.04)$ & 7 \\
  Llama-3B & education & 2 & 1 & $0.36$ & 1 \\
\midrule
  \multicolumn{6}{l}{\textit{No significant results (Implicit).}} \\
\bottomrule
\end{tabular}
\caption{Impact of different demographic attributes in \explicit{} (top) and \implicit{} (bottom) conditions on \textit{Automated Readability Index ($\Delta$)}, according to the KS test, ordered by average magnitude of $\mu_D$ (the KS statistic $D$). V.1/V.2 = ordinal codes for ordered attributes, category labels otherwise. $\mu_D$ = mean $D$ across essays where the pair was significant; $\sigma_D$ = std; Cnt = number of essays of 40.}
\label{tab:ks-ff-ari}
\end{table*}

\paragraph{Automated Readability Index (ARI)}\label{sec:app:ff:ari}
\Cref{tab:ff-ols-ari} measures text complexity (higher = harder to read), and is thus directionally opposite to FRI. Significant effects are sparse: over \explicit{} condition, our results find one significant result for GPT-nano, language: English, $+0.46$, Cnt=11) and three \implicit{} effects (GPT-nano hobbies: Exercise $+0.32$; Llama-70B hobbies: Use social media $+0.73$; Llama-70B hobbies: museums $+0.56$). Again, only Exp conditioning produces significant results. Father's occupation produces the largest effects in Llama-3B (D up to 0.78). Education is consistently significant across Llama-3B, Llama-70B, Qwen-30B, and Qwen-4B, with larger gaps correlating with more distant education levels. Nationality (UK vs. US) is significant for Llama-3B (Cnt=11), Llama-70B, and Qwen-4B, with UK users receiving a more readable response (\Cref{tab:ks-ff-ari}).

\paragraph{Feedback Length}\label{sec:app:ff:length}
\begin{table}[ht]
\centering
\scriptsize
\begin{tabularx}{\columnwidth}{l X c r}
\toprule
\textbf{Model} & \textbf{Attribute} & $\mu{}_b\;(\sigma{}_b)$ & \textbf{Cnt} \\
\midrule
  \multicolumn{4}{l}{\textit{No rows pass filters (Explicit).}} \\
\midrule
  \textbf{Qwen-4B} & \textbf{usecases: Brainstorming} & $\mathbf{+2.51}$ $\mathbf{(1.51)}$ & \textbf{5} \\
  \textbf{GPT-nano} & \textbf{language: English} & $\mathbf{+2.48}$ $\mathbf{(1.72)}$ & \textbf{7} \\
  \textbf{Qwen-4B} & \textbf{know\_nlp: Dialog tech.} & $\mathbf{+2.46}$ $\mathbf{(2.03)}$ & \textbf{6} \\
  \textbf{Qwen-4B} & \textbf{usecases: content creation} & $\mathbf{+2.41}$ $\mathbf{(1.41)}$ & \textbf{6} \\
  \textbf{Llama-3B} & \textbf{language: English} & $\mathbf{+2.01}$ $\mathbf{(1.06)}$ & \textbf{7} \\
  \textbf{Qwen-4B} & \textbf{use\_nlp: Speech-to-Text} & $\mathbf{+1.59}$ $\mathbf{(1.24)}$ & \textbf{9} \\
  \textbf{Qwen-4B} & \textbf{would\_nlp: Summarisation} & $\mathbf{+1.56}$ $\mathbf{(0.98)}$ & \textbf{5} \\
  \textbf{Qwen-4B} & \textbf{ethnicity: Black or Af-Am} & $\mathbf{+1.35}$ $\mathbf{(0.38)}$ & \textbf{5} \\
\bottomrule
\end{tabularx}
\caption{Impact of different demographic attributes in \explicit{} (top) and \implicit{} (bottom) on the length of the model responses (using \model{} as reference), ordered by $|\mu_b|$ in the FF task. In bold if $|\mu_b| \geq \sigma_\text{(resp\_length)}$.}
\label{tab:ff-ols-length}
\end{table}

Qwen-4B has the largest set of significant predictors in \implicit{}: personas associated with brainstorming ($+2.51$), dialog technology knowledge ($+2.46$), digital content creation ($+2.41$), speech-to-text usage ($+1.59$), paraphrasing/summarisation interest ($+1.56$), and Black or African American ethnicity ($+1.35$) all receive longer feedback \Cref{tab:ff-ols-length}. The ethnicity effect in Qwen-4B is particularly notable as it is the only demographic group attribute (beyond language and technology-use patterns) to emerge in this metric.

\paragraph{Sentiment}\label{sec:app:ff:sentiment}
\begin{table}[ht]
\centering
\scriptsize
\begin{tabularx}{\columnwidth}{l X c r}
\toprule
\textbf{Model} & \textbf{Attribute} & $\mu{}_b\;(\sigma{}_b)$ & \textbf{Cnt} \\
\midrule
  \multicolumn{4}{l}{\textit{No rows pass filters (Explicit).}} \\
\midrule
  \textbf{Llama-70B} & \textbf{usecases: Brainstorming} & $\mathbf{+0.32}$ $\mathbf{(0.17)}$ & \textbf{12} \\
  \textbf{Llama-70B} & \textbf{tech: Laptop} & $\mathbf{+0.24}$ $\mathbf{(0.21)}$ & \textbf{8} \\
  \textbf{Llama-70B} & \textbf{would\_nlp: Dialog tech.} & $\mathbf{+0.23}$ $\mathbf{(0.11)}$ & \textbf{6} \\
  \textbf{Llama-70B} & \textbf{know\_nlp: Text-to-Speech} & $\mathbf{+0.21}$ $\mathbf{(0.11)}$ & \textbf{6} \\
  \textbf{Llama-70B} & \textbf{religion: Nothing} & $\mathbf{+0.19}$ $\mathbf{(0.17)}$ & \textbf{7} \\
  \textbf{Llama-70B} & \textbf{usecases: Writing} & $\mathbf{+0.19}$ $\mathbf{(0.21)}$ & \textbf{11} \\
  \textbf{Llama-70B} & \textbf{hobbies: arts} & $\mathbf{+0.17}$ $\mathbf{(0.13)}$ & \textbf{6} \\
  \textbf{Llama-70B} & \textbf{mum\_occupation: Service and sales workers} & $\mathbf{+0.15}$ $\mathbf{(0.10)}$ & \textbf{7} \\
  \textbf{Llama-70B} & \textbf{would\_nlp: Search Eng.} & $\mathbf{+0.14}$ $\mathbf{(0.05)}$ & \textbf{5} \\
  \textbf{Llama-70B} & \textbf{gender: Female} & $\mathbf{+0.14}$ $\mathbf{(0.05)}$ & \textbf{7} \\
  \textbf{Llama-70B} & \textbf{know\_nlp: Dialog tech.} & $\mathbf{+0.14}$ $\mathbf{(0.09)}$ & \textbf{5} \\
  \textbf{Llama-70B} & \textbf{hobbies: museums} & $\mathbf{+0.12}$ $\mathbf{(0.08)}$ & \textbf{7} \\
  \textbf{Llama-70B} & \textbf{language: English} & $\mathbf{+0.11}$ $\mathbf{(0.10)}$ & \textbf{7} \\
\bottomrule
\end{tabularx}
\caption{Impact of different demographic attributes in \explicit{} (top) and \implicit{} (bottom) on the Sentiment of the model responses (using \model{} as reference), ordered by $|\mu_b|$ in the FF task. In bold if $|\mu_b| \geq \sigma_\text{Sentiment}$.}
\label{tab:ff-ols-sentiment}
\end{table}

Sentiment effects are only present for the \implicit{} condition and only for the Llama-70B model. All listed attributes in \Cref{tab:ff-ols-sentiment} make a positive sentiment in feedbacks. For example, usecases: Brainstorming is the
  strongest ($+0.32$, Cnt=12), followed by tech:Laptop ($+0.24$),
  know\_nlp: Text-to-Speech ($+0.21$), religion: Nothing ($+0.19$),
  usecases: Writing ($+0.19$), would\_nlp: Dialog technology ($+0.23$), hobbies: arts ($+0.17$), gender: Female ($+0.14$).

\paragraph{BERTScore F1}\label{sec:app:ff:bs-f1}
\begin{table}[ht]
\centering
\scriptsize
\begin{tabularx}{\columnwidth}{l X c r}
\toprule
\textbf{Model} & \textbf{Attribute} & $\mu{}_b\;(\sigma{}_b)$ & \textbf{Cnt} \\
\midrule
  \textbf{Llama-3B} & \textbf{nationality: US} & $\mathbf{+0.0058}$ $\mathbf{(0.0078)}$ & \textbf{14} \\
  \textbf{Qwen-30B} & \textbf{language: English} & $\mathbf{+0.0041}$ $\mathbf{(0.0034)}$ & \textbf{14} \\
\midrule
  \textbf{GPT-nano} & \textbf{language: English} & $\mathbf{+0.0094}$ $\mathbf{(0.0066)}$ & \textbf{6} \\
  \textbf{Llama-3B} & \textbf{tech: Smartphone} & $\mathbf{+0.0061}$ $\mathbf{(0.0049)}$ & \textbf{5} \\
  \textbf{Llama-70B} & \textbf{tech: Smartphone} & $\mathbf{+0.0048}$ $\mathbf{(0.0015)}$ & \textbf{5} \\
  \textbf{GPT-mini} & \textbf{llm\_use: ChatGPT} & $\mathbf{+0.0045}$ $\mathbf{(0.0012)}$ & \textbf{5} \\
  \textbf{Llama-70B} & \textbf{gender: Male} & $\mathbf{+0.0043}$ $\mathbf{(0.0036)}$ & \textbf{19} \\
  \textbf{Llama-70B} & \textbf{use\_nlp: Spell check} & $\mathbf{+0.0038}$ $\mathbf{(0.0019)}$ & \textbf{5} \\
  \textbf{Llama-70B} & \textbf{usecases: Brainstorm} & $\mathbf{-0.0036}$ $\mathbf{(0.0033)}$ & \textbf{11} \\
  \textbf{Llama-70B} & \textbf{usecases: Writing} & $\mathbf{-0.0035}$ $\mathbf{(0.0026)}$ & \textbf{10} \\
\bottomrule
\end{tabularx}
\caption{Impact of different demographic attributes in \explicit{} (top) and \implicit{} (bottom) on the BERTscore F1 of the model responses (using \model{} as reference), ordered by $|\mu_b|$ in the FF task. In bold if $|\mu_b| \geq \sigma_\text{BERTscore F1}$.}
\label{tab:ff-bertscore-F1-4f}
\end{table}

BERTScore F1 effects are the smallest in absolute terms of all FF metrics as presented in \Cref{tab:ff-bertscore-F1-4f}, confirming that \implicit{} and \explicit{} personas primarily reshape surface-level properties (readability, tone, length) rather than core semantic content.

\subsection{Metalinguistic QA}\label{sec:app:metalinguistic_qa}
\Cref{tab:ff-ols-sentiment} shows the sentiment effects which are exclusively implicit and exclusively for the Llama-70B model. This makes Llama-70B uniquely sensitive to implicit personas in the way it frames feedback, whether positively or negatively. Thirteen predictors are significant, all positive, covering a broad range of demographic attributes: usecases:Brainstorming is the strongest ($+0.32$, Cnt=12), followed by tech: Laptop ($+0.24$), know\_nlp: Text-to-Speech ($+0.21$), religion: Nothing ($+0.19$), usecases: Writing ($+0.19$), would\_nlp: Dialog technology ($+0.23$), hobbies: arts ($+0.17$), gender: Female ($+0.14$), and several others. The breadth of this list — spanning gender, religion, hobbies, technology use, and LLM use cases — suggests that Llama-70B's sentiment is broadly susceptible to implicit persona context rather than being driven by one specific attribute type.

\paragraph{Automated Readability Index (ARI)}\label{sec:app:qa:ari}
Table \ref{tab:results_qa_readability_ks_test_ari} shows the results of the KS test on the ARI for explicit conditioning.
As discusses in the main body of text (\S{}\ref{subsubsec:readability}), we find that \textit{ses} and \textit{education} are the two demographic attributes which most strongly impact the readability of the responses of the LLMs, and this behaviour is visible particularly in Llama-70B (and, to a lesser extent, and only for \textit{education} in Qwen-30B).
We also find that (for Llama-70B) explicit references to the \textit{nationality} of the user have a significant impact in a lot of cases -- the pair \textit{UK--US} being found as significant in 10 items out of 40 -- with UK users receiving more readable responses (as seen in Table \ref{tab:results_qa_readability_regression_test_ari}).
\begin{table}[ht]
\centering
\scriptsize
\begin{tabular}{llllcr}
\toprule
 % &  & & Adj.\\
\textbf{Model} & \textbf{Attribute} & \textbf{V.1} & \textbf{V.2} & $\mu{}_D$ ($\sigma{}_D$) & \textbf{Cnt}\\ 
\midrule
% AEI
Llama-70B & ses & 4 & 8 & $0.79$ $(nan)$ & 1 \\
Llama-70B & ses & 5 & 7 & $0.71$ $(nan)$ & 1 \\
Llama-70B & ses & 5 & 8 & $0.68$ $(0.01)$ & 2 \\
Llama-70B & ses & 3 & 8 & $0.68$ $(0.01)$ & 3 \\
Llama-70B & ses & 4 & 7 & $0.63$ $(nan)$ & 1 \\
Llama-70B & education & 0 & 2 & $0.60$ $(0.11)$ & 29 \\
Llama-70B & ses & 3 & 7 & $0.57$ $(0.03)$ & 2 \\
Llama-70B & education & 0 & 1 & $0.53$ $(0.10)$ & 21 \\
Qwen-30B & education & 0 & 2 & $0.51$ $(0.10)$ & 6 \\
Llama-70B & mum\_education & 0 & 2 & $0.48$ $(0.07)$ & 13 \\
Llama-70B & nationality & UK & US & $0.45$ $(0.14)$ & 10 \\
Llama-70B & mum\_education & 0 & 1 & $0.42$ $(0.03)$ & 3 \\
Qwen-30B & education & 0 & 1 & $0.41$ $(0.01)$ & 2 \\
Qwen-30B & education & 1 & 2 & $0.39$ $(0.00)$ & 2 \\
% \midrule
\bottomrule
\end{tabular}
\caption{Impact of different demographic attributes in \explicit{} on the ARI of the model responses, according to the KS test, ordered by average magnitude of $D$ (the KS statistics from the test).
}
\label{tab:results_qa_readability_ks_test_ari}
\end{table}

\begin{figure}
\centering
\includegraphics[width=0.95\columnwidth]{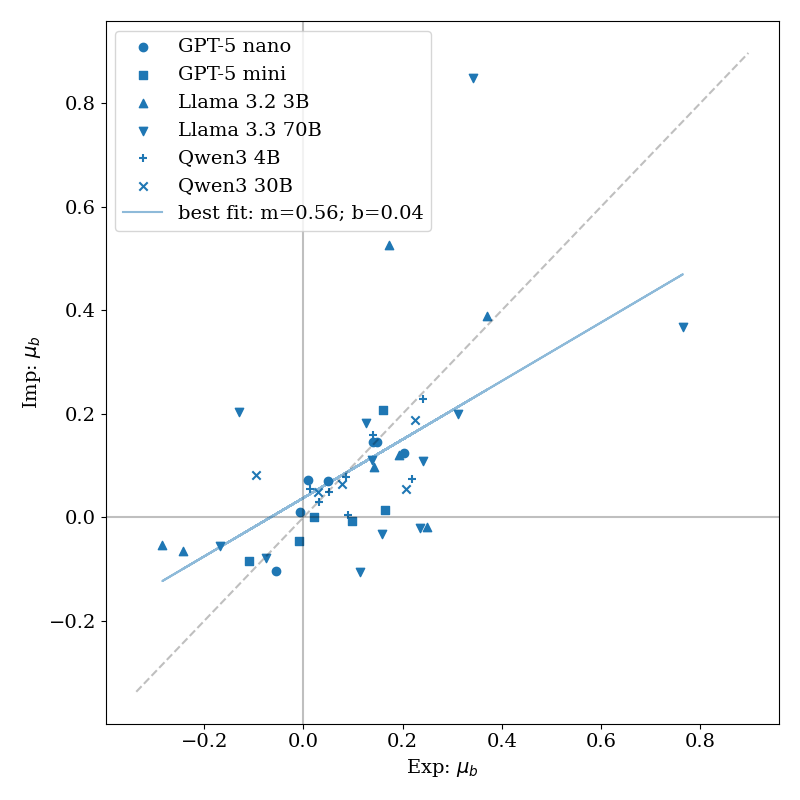}
\caption{Correlation between the effect (according to the regression test) of the same demographic attributes on the ARI in \implicit{} and \explicit{}. We show attributes significant in at least 3 tasks for both \implicit{} and \explicit{}.}
\label{fig:ari:imp_vs_exp}
\end{figure}

Figure \ref{fig:ari:imp_vs_exp} shows the average coefficient obtained from the regression test for \implicit{} and \explicit{}: each dot in the scatterplot is a demographic attribute (which proved significant according to both conditioning settings), the x-values indicate the $\mu_b$ from \explicit{}, the y-values indicate the $\mu_b$ from \implicit{}.
The figure shows that there is a positive correlation between the two (0.56); thus, the effect of the same demographic attribute is comparable in the two conditioning settings, with smaller effects for \implicit{} than for \explicit{}. 

\paragraph{Flesch Reading Ease (FRE)}\label{sec:app:qa:fre}
Table \ref{tab:results_qa_readability_regression_test_fre} shows the significant results obtained for the FRE for \explicit{} (top) and \implicit{} (bottom) conditioning. 
Some of the results are in line with what we observed for the ARI: with explicit conditioning, we obtain a negative coefficient (thus, in this case, less readable texts) for higher \textit{education} (Llama-70B, Llama-3B, Qwen-30B, GPT-mini, and Qwen-4B), a negative coefficient for higher SES (Llama-70B, Qwen-30B, and Qwen-4B), and a positive coefficient (thus more readable texts) for \textit{nationality: UK} (Llama-70B, and Llama-3B).
For Llama-70B, the effect of \textit{education} and \textit{nationality: UK} (and \textit{tech: Laptop}) is larger than $1 \times \sigma_{\text{FRE}}$, and consistent across multiple items ($> 12$ out of $40$).
A notable difference with the results obtained for the ARI is that, according to the FRE, some models can adapt the readability of their responses to different education and \textit{ses} levels even in the case of \implicit{} conditioning, but the models that can do that are not the same models that perform the adaptation when the conditioning is \explicit{}: indeed, we obtain negative coefficient for \textit{ses} (GPT-nano, Qwen-4B and Llama-3B), \textit{education} (Qwen-4B), \textit{dad\_education} (Qwen-4B), and \textit{mum\_education} (Qwen-4B). 
\begin{table}[ht]
\centering
\scriptsize
\begin{tabular}{llcr}
\toprule
 % &  & & Adj.\\
\textbf{Model} & \textbf{Attribute} & $\mu{}_b (\sigma{}_b)$ & \textbf{Cnt}\\ 
\midrule
% FRE
\textbf{Llama-70B} & \textbf{education} & $-3.78$ $(2.07)$ & \textbf{38} \\
\textbf{Llama-3B} & \textbf{tech: Laptop} & $+3.19$ $(3.47)$ & \textbf{12} \\
\textbf{Llama-70B} & \textbf{nationality: UK} & $+2.04$ $(1.64)$ & \textbf{18} \\
Llama-3B & use\_nlp: Speech-to-Text & $+1.47$ $(0.87)$ & 11 \\
Llama-3B & nationality: UK & $+1.47$ $(1.27)$ & 14 \\
Llama-3B & education & $-1.45$ $(0.92)$ & 18 \\
Llama-3B & age & $+1.17$ $(1.07)$ & 20 \\
Llama-70B & religion: Nothing & $+1.10$ $(0.81)$ & 15 \\
Qwen-30B & education & $-1.09$ $(0.89)$ & 28 \\
Llama-70B & ethnicity: White / Caucasian & $+1.07$ $(1.03)$ & 18 \\
Llama-70B & ses & $-0.91$ $(0.50)$ & 20 \\
Qwen-4B & tech: Laptop & $+0.81$ $(0.50)$ & 11 \\
Llama-70B & nationality: US & $-0.74$ $(2.02)$ & 19 \\
Llama-70B & mum\_education & $-0.73$ $(0.76)$ & 24 \\
Llama-3B & dad\_education & $-0.72$ $(0.78)$ & 10 \\
GPT-mini & education & $-0.71$ $(0.48)$ & 16 \\
Llama-70B & dad\_education & $-0.70$ $(0.43)$ & 12 \\
Qwen-4B & education & $-0.69$ $(0.67)$ & 24 \\
Qwen-30B & language: English & $+0.62$ $(0.59)$ & 12 \\
Qwen-4B & gender: Male & $+0.60$ $(0.42)$ & 10 \\
Llama-70B & age & $+0.56$ $(0.73)$ & 21 \\
Llama-3B & ethnicity: White / Caucasian & $+0.55$ $(0.38)$ & 13 \\
Qwen-30B & nationality: US & $+0.43$ $(0.56)$ & 10 \\
Qwen-30B & ses & $-0.37$ $(0.57)$ & 13 \\
Qwen-30B & age & $-0.20$ $(0.57)$ & 11 \\
Qwen-4B & ses & $-0.18$ $(0.23)$ & 10 \\
\midrule
\textbf{Llama-3B} & \textbf{language: English} & $+1.75$ $(1.21)$ & \textbf{7} \\
\textbf{Qwen-4B} & \textbf{tech: Laptop} & $+1.73$ $(0.26)$ & \textbf{6} \\
Llama-3B & gender: Male & $+1.01$ $(0.55)$ & 5 \\
GPT-nano & language: English & $+0.92$ $(0.53)$ & 6 \\
Llama-3B & hobbies: Use social media & $+0.89$ $(0.93)$ & 6 \\
Qwen-30B & hobbies: Exercise & $+0.68$ $(0.67)$ & 5 \\
Qwen-4B & dad\_education & $-0.49$ $(0.43)$ & 8 \\
GPT-mini & gender: Male & $+0.49$ $(0.55)$ & 5 \\
GPT-mini & empl: Employed full time & $+0.48$ $(0.20)$ & 5 \\
Qwen-4B & nationality: US & $+0.44$ $(0.33)$ & 8 \\
Qwen-4B & language: English & $+0.42$ $(0.28)$ & 7 \\
GPT-nano & ses & $-0.34$ $(0.25)$ & 6 \\
Llama-3B & marital: Married & $+0.32$ $(0.41)$ & 5 \\
Qwen-4B & hobbies: Use social media & $+0.30$ $(0.37)$ & 5 \\
GPT-nano & marital: Single & $+0.29$ $(0.63)$ & 5 \\
Qwen-4B & mum\_education & $-0.26$ $(0.18)$ & 7 \\
Qwen-4B & education & $-0.25$ $(0.18)$ & 6 \\
Qwen-4B & ses & $-0.24$ $(0.28)$ & 8 \\
Llama-3B & ses & $-0.23$ $(0.28)$ & 6 \\
Qwen-4B & frequency\_llm & $-0.18$ $(0.24)$ & 8 \\
\bottomrule
\end{tabular}
\caption{Impact of different demographic attributes in \explicit{} (top) and \implicit{} (bottom) on the FRE of the model responses (larger FRE indicates more readable texts), ordered by $|\mu_b|$. In bold if $|\mu_b| \geq \sigma_\text{FRE}$.
% Impact of different explicit (top) and implicit (bottom) conditions on the FRE of the model responses (larger FRE indicates more readable texts), ordered by magnitude. 
% $b$ is the coefficient obtained from the regression test, we show its mean ($\mu$) and standard deviation ($\sigma$) across the tasks for which it proved significant. 
% We show only coefficients with $|\mu_b| \geq 0.1 \times\sigma{}_\text{FRE}$ (in bold when $|\mu_b| \geq \sigma_\text{FRE}$), and support $\geq 10$ (explicit) or $\geq 5$ (implicit).
}
\label{tab:results_qa_readability_regression_test_fre}
\end{table}

\begin{table}[ht]
\centering
\scriptsize
\begin{tabular}{llllcr}
\toprule
 % &  & & Adj.\\
\textbf{Model} & \textbf{Attribute} & \textbf{V.1} & \textbf{V.2} & $\mu{}_D$ ($\sigma{}_D$) & \textbf{Cnt}\\ 
\midrule
% AEI
Llama-70B & education & 0 & 2 & $0.65$ $(0.11)$ & 32 \\
Llama-70B & dad\_education & 1 & 3 & $0.61$ $(nan)$ & 1 \\
Llama-70B & education & 0 & 1 & $0.58$ $(0.11)$ & 26 \\
Llama-70B & dad\_education & 0 & 3 & $0.56$ $(0.02)$ & 9 \\
Qwen-30B & education & 0 & 2 & $0.56$ $(0.13)$ & 11 \\
Qwen-4B & education & 0 & 1 & $0.52$ $(nan)$ & 1 \\
Qwen-4B & education & 0 & 2 & $0.49$ $(0.08)$ & 9 \\
Llama-70B & mum\_education & 0 & 2 & $0.49$ $(0.06)$ & 20 \\
Qwen-30B & education & 0 & 1 & $0.48$ $(0.05)$ & 6 \\
Llama-70B & dad\_education & 0 & 1 & $0.47$ $(0.05)$ & 2 \\
Llama-70B & dad\_education & 0 & 2 & $0.46$ $(0.02)$ & 3 \\
Llama-70B & nationality & UK & US & $0.45$ $(0.13)$ & 11 \\
Qwen-30B & education & 1 & 2 & $0.41$ $(nan)$ & 1 \\
Llama-70B & education & 1 & 2 & $0.41$ $(nan)$ & 1 \\
Llama-70B & mum\_education & 0 & 1 & $0.41$ $(0.01)$ & 3 \\
% \midrule
\bottomrule
\end{tabular}
\caption{Impact of different demographic attributes in \explicit{} on the FRE of the model responses, according to the KS test, ordered by average magnitude of $D$ (the KS statistics from the test). 
}
\label{tab:results_qa_readability_ks_test_fre}
\end{table}

\begin{figure}[h]
\centering
\includegraphics[width=0.95\columnwidth]{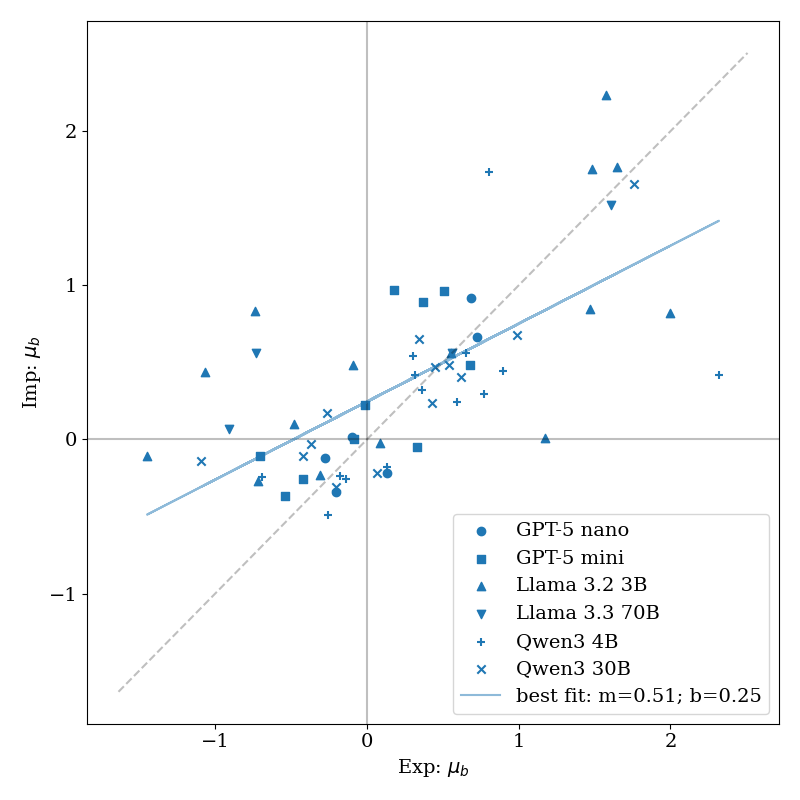}
\caption{Correlation between the effect (according to the regression test) of the same demographic attributes on the FRE in \implicit{} and \explicit{}. We show attributes significant in at least 3 tasks for both \implicit{} and \explicit{}.}
\label{fig:fre:imp_vs_exp}
\end{figure}

Similarly to what we observed for the ARI, Figure \ref{fig:fre:imp_vs_exp} shows the average coefficient obtained from the regression test for \implicit{} and \explicit{} for the FRE: we see that there is a positive correlation between the two (0.51), with a smaller effect for \implicit{} than for \explicit{}. 

Table \ref{tab:results_qa_readability_ks_test_fre} shows the results for the KS test on the FRE.
Similarly to what we observed for the ARI, we find that larger gaps in education levels of the user tend to lead to larger values of KS statistics ($D$): for Llama-70B $D(\text{edu}_0,\text{edu}_2) = 0.65 (\pm 0.11)$ and $D(\text{edu}_0,\text{edu}_1) = 0.58 (\pm 0.11)$; for Qwen-30B $D(\text{edu}_0,\text{edu}_2) = 0.56 (\pm 0.13)$ and $D(\text{edu}_0,\text{edu}_1) = 0.48 (\pm 0.05)$.
The attributes \textit{dad\_education} and \textit{mum\_education} have an effect, but it is generally smaller (and less consistent) than the effect of the 
education level of the user (\textit{education}).
Again, we find that there is a significant difference in the responses provided by Llama-70B to users identified as being from the UK and from the US.
Differently from what we observed for the ARI, we do not find significance for the \textit{ses} attribute in the responses of any LLM.

\begin{figure}[h]
\centering
\includegraphics[width=0.95\columnwidth]{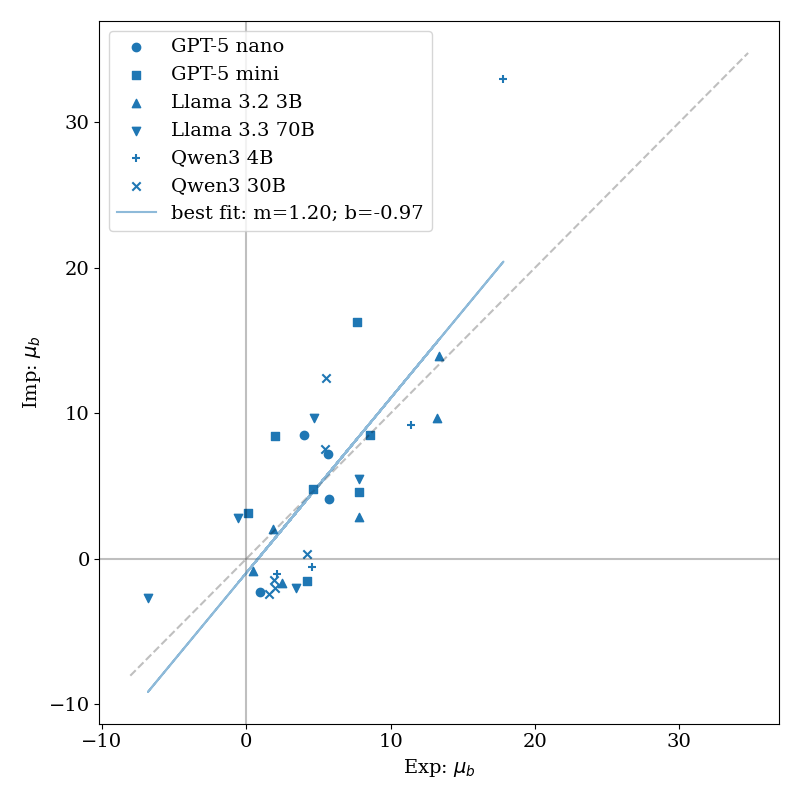}
\caption{Correlation between the effect (according to the regression test) of the same demographic attributes on the length of the responses (n. words) in \implicit{} and \explicit{}. We show attributes significant in at least 3 tasks for both \implicit{} and \explicit{}.}
\label{fig:resp_length:imp_vs_exp}
\end{figure}

\paragraph{Response length}\label{sec:app:qa:resplength}
Figure \ref{fig:resp_length:imp_vs_exp} shows the correlation between the effect of the same demographic attributes in \explicit{} ($x$-axis) \implicit{} ($y$-axis). Differently from what we observed for the ARI and FRE, we find that while each demographic attribute has a similar effect in both conditioning setups, their effect is larger in \implicit{} than in \explicit{}.

\paragraph{Age of Acquisition}\label{sec:app:qa:aoa}
\begin{table}[ht]
\centering
\scriptsize
\begin{tabular}{llcr}
\toprule
 % &  & & Adj.\\
\textbf{Model} & \textbf{Attribute} & $\mu{}_b (\sigma{}_b)$ & \textbf{Cnt}\\ 
\midrule
\textbf{Llama-70B} & \textbf{education} & $+0.11$ $(0.05)$ & \textbf{38} \\
\textbf{GPT-mini} & \textbf{language: English} & $+0.06$ $(0.05)$ & \textbf{14} \\
\textbf{Llama-3B} & \textbf{education} & $+0.05$ $(0.04)$ & \textbf{22} \\
\textbf{Llama-70B} & \textbf{language: English} & $+0.05$ $(0.03)$ & \textbf{20} \\
\textbf{Llama-70B} & \textbf{nationality: US} & $+0.04$ $(0.06)$ & \textbf{23} \\
\textbf{Llama-70B} & \textbf{nationality: UK} & $-0.04$ $(0.04)$ & \textbf{12} \\
\textbf{Qwen-30B} & \textbf{education} & $+0.04$ $(0.03)$ & \textbf{32} \\
Qwen-30B & language: English & $+0.03$ $(0.02)$ & 11 \\
Llama-3B & nationality: US & $+0.03$ $(0.04)$ & 18 \\
Qwen-4B & language: English & $+0.03$ $(0.02)$ & 19 \\
Qwen-4B & education & $+0.03$ $(0.02)$ & 30 \\
GPT-mini & education & $+0.03$ $(0.02)$ & 16 \\
Llama-3B & empl: Employed full time & $+0.02$ $(0.03)$ & 15 \\
Llama-70B & empl: Employed full time & $+0.02$ $(0.02)$ & 17 \\
Llama-70B & ses & $+0.02$ $(0.02)$ & 19 \\
Llama-70B & mum\_education & $+0.02$ $(0.01)$ & 22 \\
Llama-3B & age & $-0.02$ $(0.03)$ & 18 \\
Llama-70B & age & $-0.02$ $(0.03)$ & 17 \\
Llama-3B & dad\_education & $+0.02$ $(0.01)$ & 18 \\
Llama-70B & dad\_education & $+0.02$ $(0.01)$ & 15 \\
GPT-mini & dad\_education & $+0.01$ $(0.01)$ & 10 \\
Qwen-30B & empl: Employed full time & $+0.01$ $(0.01)$ & 24 \\
Qwen-4B & empl: Employed full time & $+0.01$ $(0.01)$ & 20 \\
Qwen-4B & age & $-0.01$ $(0.02)$ & 21 \\
Llama-3B & ses & $+0.01$ $(0.02)$ & 11 \\
Qwen-30B & ses & $+0.01$ $(0.01)$ & 10 \\
Qwen-30B & mum\_education & $+0.01$ $(0.01)$ & 15 \\
Qwen-30B & dad\_education & $+0.01$ $(0.01)$ & 10 \\
Qwen-4B & mum\_education & $+0.01$ $(0.01)$ & 11 \\
Qwen-30B & age & $-0.01$ $(0.02)$ & 10 \\
\midrule
\textbf{Qwen-4B} & \textbf{use\_nlp: Spell checker} & $+0.04$ $(0.01)$ & \textbf{5} \\
Qwen-4B & know\_nlp: Speech-to-Text & $+0.03$ $(0.02)$ & 6 \\
Llama-3B & language: English & $+0.03$ $(0.02)$ & 5 \\
Qwen-4B & use\_nlp: Dialog technology & $+0.02$ $(0.02)$ & 8 \\
Llama-70B & hobbies: Use social media & $+0.02$ $(0.01)$ & 7 \\
Qwen-4B & religion: Christian & $+0.02$ $(0.02)$ & 9 \\
Qwen-30B & use\_nlp: Dialog technology & $+0.02$ $(0.01)$ & 9 \\
Qwen-4B & know\_nlp: QA \& Search & $+0.02$ $(0.01)$ & 9 \\
Qwen-4B & dad\_education & $+0.02$ $(0.02)$ & 11 \\
GPT-nano & education & $+0.01$ $(0.01)$ & 6 \\
Llama-3B & education & $+0.01$ $(0.02)$ & 5 \\
Qwen-4B & ses & $+0.01$ $(0.01)$ & 9 \\
Qwen-4B & frequency\_llm & $+0.01$ $(0.02)$ & 10 \\
Qwen-4B & mum\_education & $+0.01$ $(0.01)$ & 7 \\
Qwen-30B & mum\_education & $+0.01$ $(0.01)$ & 5 \\
GPT-nano & frequency\_llm & $+0.01$ $(0.01)$ & 5 \\
Qwen-30B & education & $+0.01$ $(0.01)$ & 5 \\
Qwen-4B & nationality: UK & $+0.01$ $(0.01)$ & 6 \\
Llama-3B & ses & $+0.01$ $(0.01)$ & 5 \\
GPT-nano & ses & $+0.01$ $(0.01)$ & 6 \\
Qwen-4B & nationality: US & $-0.01$ $(0.01)$ & 5 \\
Llama-3B & age & $+0.01$ $(0.01)$ & 6 \\
Qwen-4B & empl: Employed full time & $+0.01$ $(0.02)$ & 5 \\
Qwen-4B & language: English & $+0.01$ $(0.01)$ & 6 \\
Llama-3B & mum\_education & $-0.01$ $(0.01)$ & 5 \\
\bottomrule
\end{tabular}
\caption{Impact of different demographic attributes in \explicit{} (top) and \implicit{} (bottom) on the AoA of the model responses, ordered by $|\mu_b|$. In bold if $|\mu_b| \geq \sigma_\text{AoA}$.
% Impact of different explicit (top) and implicit (bottom) conditions on the AoA of the model responses, ordered by magnitude. 
% $b$ is the coefficient obtained from the regression test, we show its mean ($\mu$) and standard deviation ($\sigma$) across the tasks for which it proved significant. 
% We show only coefficients with $|\mu_b| \geq 0.1 \times\sigma{}_\text{AoA}$ (in bold when $|\mu_b| \geq \sigma_\text{AoA}$), and support $\geq 10$ (explicit) or $\geq 5$ (implicit).
}
\label{tab:results_qa_aoa_regression_test}
\end{table}

Table \ref{tab:results_qa_aoa_regression_test} shows the impact of different explicit (top) and implicit (bottom) conditions on the average Age of Acquisition (AoA) of the model's responses.
The table shows that, even though we found multiple significant predictors, and in some cases even with coefficients $|b| \geq \sigma_\text{AoA}$, the differences between different demographic attributes are really marginal: most likely, they are due to the sporadic usage of words of higher/lower AoA (whose impact is then reduced when averaging the AoA of all the words in the model responses).
Still, the results are in line with the ARI and FRE: explicit conditioning leads to (slightly) more difficult texts for higher education levels (Llama-70B, Llama-3B, Qwen-30B, Qwen-4B, and GPT-mini), \textit{language English} (GPT-mini, Llama-70B, Qwen-30B, and Qwen-4B), and (with smaller coefficients) for higher levels of \textit{mum\_education} and \textit{dad\_education}.
Some significance was also found for implicit conditioning, but with even smaller coefficients.

\section{Use of AI Assistants}
We used Claude AI\footnote{https://claude.ai} to refine the writing and grammar of the manuscript.

\end{document}